%% file: acl_latex.tex
\pdfoutput=1

\documentclass[11pt]{article}

\usepackage[final]{acl}

\usepackage{times}
\usepackage{latexsym}

\usepackage[T1]{fontenc}

\usepackage[utf8]{inputenc}

\usepackage{microtype}

\usepackage{inconsolata}

\usepackage{graphicx}

\newcommand{\cellgrey}[1]{\cellcolor[HTML]{E8E8E8}{#1}}

\usepackage{multirow}
\usepackage{makecell}
\usepackage{booktabs}
\usepackage{pifont}
\usepackage{amsmath}
\usepackage{float}
\usepackage{graphicx}
\usepackage{float}  
\usepackage{subfigure}  
\usepackage{amsmath}
\usepackage{caption}
\usepackage{float}  
\usepackage{amssymb}
\usepackage{amsmath}
\usepackage{algorithm}
\usepackage{algpseudocode}
\usepackage{colortbl}

\usepackage{tcolorbox}
\usepackage{xcolor} 
\definecolor{grenn_1}{RGB}{143, 255, 221} 
\definecolor{grenn_2}{RGB}{88, 201, 185} 

\usepackage[frozencache,cachedir=minted-cache]{minted}

\tcbuselibrary{skins, breakable}
\usepackage{hyperref}

\newcounter{mybox}

\usepackage{longtable}
\usepackage{tcolorbox}
\usepackage{ulem}
\usepackage{listings}

\tcbuselibrary{listingsutf8}
\definecolor{delim}{RGB}{20,105,176}
\definecolor{numb}{RGB}{106, 109, 32}
\definecolor{string}{rgb}{0.64,0.08,0.08}
\lstdefinestyle{dialogue}{
    frame=tb,
    basicstyle=\ttfamily,
    columns=fullflexible,
    breaklines=true,
    postbreak=\mbox{\textcolor{red}{$\hookrightarrow$}\space},
    breakindent=1em,
    xleftmargin=2em,
    xrightmargin=2em,       
    backgroundcolor=\color{blue!5}, 
    frame=single, 
    rulecolor=\color{black}, 
    basicstyle=\large, 
    columns=fullflexible, 
    numbers=none, 
    stepnumber=1, 
    numbersep=5pt, 
    tabsize=4, 
    breaklines=true 
}

\lstdefinelanguage{json}{
    frame=tb,
    rulecolor=\color{black},
    showspaces=false,
    showtabs=false,
    breaklines=true,
    postbreak=\raisebox{0ex}[0ex][0ex]{\ensuremath{\color{gray}\hookrightarrow\space}},
    breakatwhitespace=true,
    basicstyle=\ttfamily\small,
    upquote=true,
    morestring=[b]",
    stringstyle=\color{string},
    literate=
     *{0}{{{\color{numb}0}}}{1}
      {1}{{{\color{numb}1}}}{1}
      {2}{{{\color{numb}2}}}{1}
      {3}{{{\color{numb}3}}}{1}
      {4}{{{\color{numb}4}}}{1}
      {5}{{{\color{numb}5}}}{1}
      {6}{{{\color{numb}6}}}{1}
      {7}{{{\color{numb}7}}}{1}
      {8}{{{\color{numb}8}}}{1}
      {9}{{{\color{numb}9}}}{1}
      {\{}{{{\color{delim}{\{}}}}{1}
      {\}}{{{\color{delim}{\}}}}}{1}
      {[}{{{\color{delim}{[}}}}{1}
      {]}{{{\color{delim}{]}}}}{1},
    backgroundcolor=\color{yellow!5}, 
    frame=single, 
    rulecolor=\color{black}, 
    basicstyle=\small, 
    columns=fullflexible, 
    numbers=none, 
    stepnumber=1, 
    numbersep=5pt, 
    tabsize=2, 
    breaklines=true 
}

\title{ChatSOP: An SOP-Guided MCTS Planning Framework \\for Controllable LLM Dialogue Agents}
\author{
  \textbf{Zhigen Li}\textsuperscript{1,2}, \textbf{Jianxiang Peng}\textsuperscript{1}, \textbf{Yanmeng Wang}\textsuperscript{2}, \textbf{Yong Cao}\textsuperscript{3}, \textbf{Tianhao Shen}\textsuperscript{1} \\
  \textbf{Minghui Zhang}\textsuperscript{1}, \textbf{Linxi Su}\textsuperscript{1}, \textbf{Shang Wu}\textsuperscript{4}, \textbf{Yihang Wu}\textsuperscript{1}, \textbf{Yuqian Wang}\textsuperscript{1} \\
  \textbf{Ye Wang}\textsuperscript{2}, \textbf{Wei Hu}\textsuperscript{2}, \textbf{Jianfeng Li}\textsuperscript{2}, \textbf{Shaojun Wang}\textsuperscript{2}, \textbf{Jing Xiao}\textsuperscript{2} \\
  \textbf{Deyi Xiong}\textsuperscript{1}\thanks{Corresponding author} \\
  \textsuperscript{1}TJUNLP Lab, College of Intelligence and Computing, Tianjin University \\
  \textsuperscript{2}Ping An Technology \quad \textsuperscript{3}Tübingen AI Center, University of Tübingen \\
  \textsuperscript{4}Kunming University of Science and Technology \\
  \texttt{\{zhigenli, dyxiong\}@tju.edu.cn}
}

\usepackage{fancyhdr}
\usepackage{lipsum} 
\fancypagestyle{firstpage}{
    \fancyfoot[C]{\footnotesize
    \makebox[0pt][c]{\textit{Proceedings of the 63rd Annual Meeting of the Association for Computational Linguistics (Volume 1: Long Papers)}, pages 17637--17659}\\
    July 27 - August 1, 2025 \textcopyright{} 2025 Association for Computational Linguistics}
}

\begin{document}

\thispagestyle{firstpage}

\maketitle
\begin{abstract}
Dialogue agents powered by Large Language Models (LLMs) show superior performance in various tasks. Despite the better user understanding and human-like responses, their \textbf{lack of controllability} remains a key challenge, often leading to unfocused conversations or task failure. 
To address this, we introduce Standard Operating Procedure (SOP) to regulate dialogue flow.
Specifically, we propose \textbf{ChatSOP}, a novel SOP-guided Monte Carlo Tree Search (MCTS) planning framework designed to enhance the controllability of LLM-driven dialogue agents. 
To enable this, we curate a dataset comprising SOP-annotated multi-scenario dialogues, generated using a semi-automated role-playing system with GPT-4o and validated through strict manual quality control. 
Additionally, we propose a novel method that integrates Chain of Thought reasoning with supervised fine-tuning for SOP prediction and utilizes SOP-guided Monte Carlo Tree Search for optimal action planning during dialogues. 
Experimental results demonstrate the effectiveness of our method, such as achieving a 27.95\% improvement in action accuracy compared to baseline models based on GPT-3.5 and also showing notable gains for open-source models.
Dataset and codes are publicly available.\footnote{\url{https://github.com/tjunlp-lab/ChatSOP}} 
\end{abstract}

\section{Introduction}
Task-oriented dialogue agents are essential for applications such as hotel booking, technical support, and customer service \citep{ouyang2022training, moradshahi2023x}. 
Recent advancements leverage Large Language Models' (LLMs) in-context learning ability to improve understanding, generate human-like responses, and adapt to diverse domains \citep{liu-etal-2024-lara, yi2024survey}. 

\begin{figure}[t]
    \vskip 1em 
    \centering
    \includegraphics[width=1.0\linewidth]{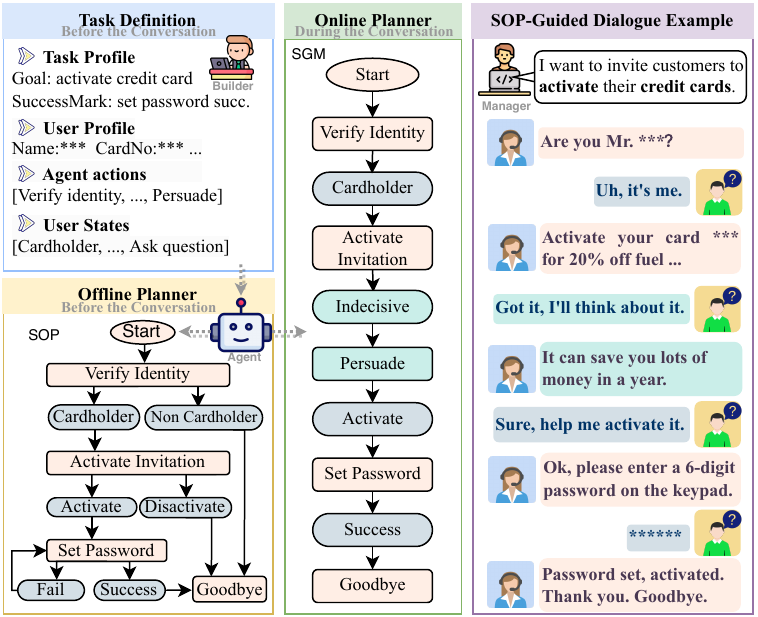}
    \caption{
    The three-stage workflow of our SOP-based planning framework for task-oriented dialogue with LLMs includes: (1) task definition, (2) the offline planner predicts the standard operating procedure (SOP) before the conversation, and (3) the online planner selects a dialogue action via SOP-guided MCTS (SGM) during the conversation. Control-sensitive actions are highlighted in orange. The right panel illustrates a dialogue in which the agent successfully activates a bank credit card by following the predicted SOP.
    }
    \label{fig: cogd_example}
    \vskip -0.5em
\end{figure}

However, despite the enhanced intelligence powered by LLMs, a key challenge that persists in current dialogue agents is the lack of controllability \cite{achiam2023gpt, hadi2024large}.
For instance, as illustrated in Figure \ref{fig: cogd_example} right panel, activating a credit card requires a specific sequence of steps—verifying personal information, creating a password, and activating the account. Omitting any step could result in task failure. Therefore, developing effective mechanisms to ensure greater control and goal-directed actions is crucial for LLM-based task-oriented dialogue agents.
To address this challenge, we introduce a Standard Operating Procedure (SOP) to regulate the dialogue flow strictly following the task process.
Specifically, we propose \textbf{ChatSOP}, an SOP-guided Monte Carlo Tree Search planning framework designed to \textbf{enhance the controllability of LLM-driven dialogue agents}. 
Unlike methods relying on manually annotated dialogue flows or training data, our approach requires only user-provided task definitions and goals, enabling autonomous planning for better generalizability at low cost.

To define procedures required by specific tasks, we first introduce a standard operating procedure (SOP) to control the dialogue states inspired by the traditional dialogue framework \cite{anantha2020open, zhu2022solving}, and construct a dataset designed for multi-scenario conversations, consisting of task descriptions, controlled SOP, and complete dialogues. This dataset is constructed through a four-step role-playing system utilizing GPT-4o, combined with human validation and modification to ensure intermediate dialogue control and data quality. To the best of our knowledge, this is the first dataset that provides SOP intermediate annotations, which could also be explored to evaluate general-purpose LLM agents.

To enable LLM dialogue agents to complete goal-driven tasks with controllability, we propose a three-stage SOP-based planning framework. 
As illustrated in Figure \ref{fig: cogd_example}, when provided with the task definition and dialogue goal, an offline planner predicts a task-specific Standard Operating Procedure before the conversation. 
During the conversation, an online planner leverages SOP-guided Monte Carlo Tree Search (SGM) to select the optimal action that not only follows the SOP but also proactively guides the user toward the dialogue goal.

Experimental results demonstrate that our method achieves significant improvements in task success rate, with a 27.95\% increase in overall action accuracy compared to baseline based on GPT-3.5. Additionally, for open-source models, larger models yield substantially better results, as evidenced by the performance gap between Llama3-70B (78.35\%) and Llama3-8B (46.85\%), highlighting their ability differences in dialogue tasks. 

In summary, we make \textbf{three main contributions}: 1) We develop a semi-automatic role-playing framework with manual review, then construct the first SOP-annotated dataset to support research on controllable dialogue agents. 2) We propose a planning-based framework integrating SOP and MCTS to enhance controllability of LLM task-oriented dialogue; 3) Extensive experiments via automatic and human evaluation demonstrate the utility of our dataset and effectiveness of our method, achieving superior performance in offline SOP prediction and online dialogue planning.

\section{Related Work}
\textbf{Dialogue Agents.} Existing approaches to dialogue agents can be categorized into four groups: conversational question answering (CQA) \citep{singhal2023towards, shi2024corecode,liu-etal-2023-tab}, open-domain dialogue (ODD) \citep{ ouyang2022training,zhang2023napg,liu-etal-2023-tab}, task-oriented dialogue (TOD) \citep{quan2020risawoz, you2024tclnet}, and conversational recommender systems (CRS) \citep{zhang2021kers, wang2023target}. CQA and ODD passively respond to users with knowledgeable or engaging conversations. TOD provides functional services following a structured process driven by training data \citep{budzianowski2018multiwoz, quan2020risawoz, moradshahi-etal-2023-x}. CRS plans dialog actions to guide conversations toward given goals \citep{wu2019proactive} but often fails to handle complex tasks requiring strict sequential constraints \citep{Akyar12, zhou2023agents}.  

\textbf{Dialogue Planning and Policy Optimization.}  Traditional dialogue planning research has focused on subgoal generation \citep{zhang2021kers}, the next round of dialogue transition strategy \citep{tang2019target}, hierarchical strategy \citep{kishinami2022target}.
While Recent frameworks explore planning dialogue paths using basic knowledge, goal-oriented dialogue planning frameworks, and proactive transitions between dialogue stages \citep{wang2022follow}.

Reinforcement Learning (RL) has long been a cornerstone for optimizing dialogue policies.
The advent of LLMs has significantly advanced this area, enabling the use of step-by-step RL for task-oriented dialogue \citep{du-etal-2024-rewarding}, the development of more proactive systems \citep{dong-etal-2025-protod}, and the introduction of a dual-process planner framework combining LLMs and MCTS for policy optimization \citep{he-etal-2024-planning}.
However, a persistent challenge, as highlighted by \citet{wang2023dialogue}, is that many approaches employ greedy single-turn prediction strategies but ignore the interdependencies of global policies, resulting in uncontrollability from the perspective of global conversation. Thus, we propose a SOP-guided planning approach to address this issue.

\textbf{Planning and Reasoning of LLMs.} LLMs show prowess in planning and reasoning. Examples include Chain-of-Thought \citep{NEURIPS2022_8bb0d291}, its variants \citep{kojima2022large} , Self-Consistency \citep{wang2022self}, Least-to-most Prompting \citep{zhou2022least} and Self-Assessment ~\citep{welleck2022generating,shinn2023reflexion}. Recent efforts have used more complex reasoning processes, offering new avenues to improve and optimize the reasoning process \citep{zhang-xiong-2025-backmath}. For example, \citet{yao2023tree} apply heuristic-based search methods, such as depth-first and breadth-first search, to discover optimized reasoning pathways. \citet{zhu2022solving} and \citet{hao2023reasoning} have introduced MCTS to reason steps for complex math or logical reasoning. Unlike them, we use MCTS for dialogue planning, encoding SOP constraints into its expansion and simulation steps.

\section{Problem Formulation} 
\label{taskdef}

In our work, we decompose the dialogue tasks into three steps: task initialization, Standard Operating Procedures (SOPs) prediction, and task execution via dialogue generation based on SOPs.

\textbf{Task Initialization.} When a user specifies a task, we collect the user-defined task profile and associated user information profile, denoted as $p$. These profiles include textual descriptions of the goal of the task, relevant background knowledge, and user-specific information.
We define a multi-turn dialogue as $D_t = \{(u_t, s_t, a_t, r_t)\}_{t=1}^T$, where each tuple $(u_t, s_t, a_t, r_t)$ denotes the $t$-th turn of the dialogue. Here, $u_t$ is a user utterance, $s_t$ represents user states, $a_t$ denotes agent actions, and $r_t$ is the agent response utterance. Please refer to Appendix \ref{apdx: task_define_example} for detailed examples.

Then, as shown in Figure \ref{fig: cogd_example}(a), we define the SOP graph $G_t$ as a directed graph, where the vertices are annotated with agent actions $a_t$ and user states $s_t$, and the edges represent the connections between these vertices. 

\textbf{SOP Prediction.}
As an intermediate step for controlling dialogue generation, SOP prediction is to predict the connections of SOP graph nodes with given user states $s_t$ and agent actions $a_t$, enabling the construction of a complete SOP graph $G$. To do so, we introduce the adjacency List $\mathcal{M}$ to represent all the connections in the SOP graph. Thus, the task is defined as follows:

\begin{equation}
    \hat{m} = \arg\max_{m \in \mathcal{M}} \; P( m \mid s_t, a_t),  
\end{equation}

It is important to emphasize that any modifications or deletions to the SOP graph will result in inaccurate task completion.

\textbf{Dialogue Generation.} The aim of this task is to first generate user states at turn $t+1$ based on profiles $p$, historical dialogue $D_t$ with the predicted SOP graph $G_t$ and user utterance $u_{t+1}$:
\begin{equation}
    \hat{s}_{t+1} = \arg\max_{s_{t+1}} \; P(s_{t+1} \mid D_t, p, u_{t+1}, \mathcal{M})
\end{equation}

Then, we predict the agent action and generate a response at turn $t+1$ as follows:
\begin{equation}
    \hat{a}_{t+1} = \arg\max_{a_{t+1}} \; P(a_{t+1} \mid D_t, p, \hat{s}_{t+1}, \mathcal{M})
\end{equation}

\begin{equation}
    \hat{r}_{t+1} = \arg\max_{r_{t+1}} \; P(r_{t+1} \mid D_t, p, u_{t+1}, \hat{a}_{t+1}, \mathcal{M})
\end{equation}

\begin{figure}[t]
    \centering
    \includegraphics[width=1.0\linewidth]{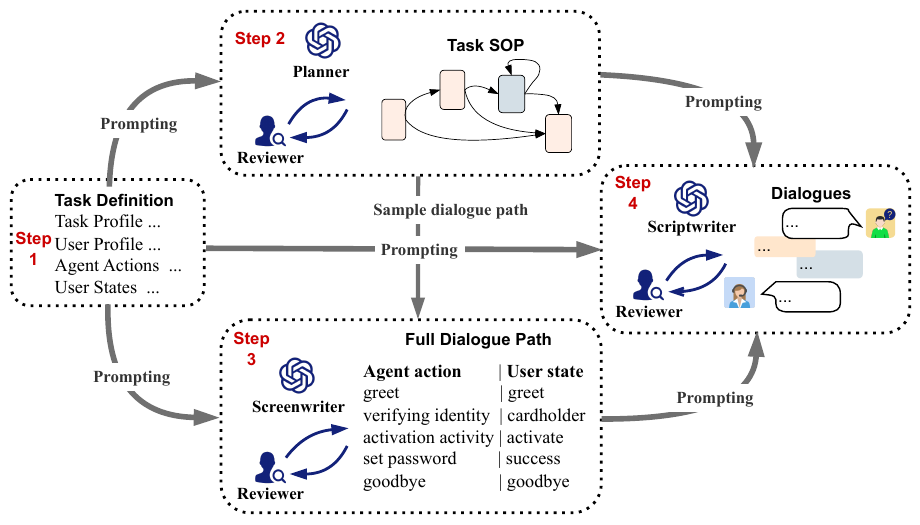}
    \vskip -0.5em
    \caption{Overview of our role-playing framework for \textit{SOPDAIL} dataset construction. This process uses human-LLM collaboration across four steps: (1) An annotator provides the initial task definition. (2) An LLM acts as a ``Planner'' to generate a candidate SOP graph, which is then reviewed by the human. (3) An LLM ``Screenwriter'' samples a path from the SOP and enriches it. (4) An LLM ``Scriptwriter'' generates the final dialogue script, which is again validated by a human reviewer. All LLMs are GPT-4o.}
    \label{fig: dataflow}
\end{figure}

\section{The ChatSOP Dataset}

Before introducing our method, we present the creation of an SOP-annotated dialogue dataset, namely \textit{SOPDAIL}, which includes designing a role-playing framework to simulate diverse scenarios, conducting human annotations to ensure high-quality, and performing a comprehensive dataset analysis.

\begin{table*}[tbp]
\centering
\scalebox{0.77}{
\begin{tabular}{lcccccccc}
\toprule
\multicolumn{1}{l}{\textbf{Dataset}} & 
\multicolumn{1}{c}{\textbf{Participants}} & 
\multicolumn{1}{c}{\textbf{Agent Goals}} &
\multicolumn{1}{c}{\textbf{PA}} &
\multicolumn{1}{c}{\textbf{CT}} & 
\multicolumn{1}{c}{\textbf{Lang}} & 
\multicolumn{1}{c}{\textbf{\#Domains}} & 
\multicolumn{1}{c}{\textbf{\#Tasks}} & 
\multicolumn{1}{c}{\textbf{\#Dialogues}} \\
\midrule
DSTC2  \citep{henderson2014second}               & Crowd    & N/A               & \textcolor{red}{\ding{55}} & \textcolor{green}{\ding{51}} & English    & 1    & 1   & 1,612 \\
DSTC4  \citep{kim2017fourth}               & Experts  & N/A               & \textcolor{red}{\ding{55}} & \textcolor{green}{\ding{51}} & English    & 1    & 1   & 35 \\
CrossWOZ  \citep{zhu2020crosswoz} & Rules,Crowd    & N/A               & \textcolor{red}{\ding{55}} & \textcolor{green}{\ding{51}} & Chinese    & 5    & 5   & 5,012 \\
SGD  \citep{rastogi2020towards}               & Rules,Crowd  & N/A               & \textcolor{red}{\ding{55}} & \textcolor{green}{\ding{51}} & English    & 16    & 26   & 16,142 \\
OTTers \citep{sevegnani2021otters}         & Crowd    & Topics            & \textcolor{green}{\ding{51}} & \textcolor{red}{\ding{55}} & English    &Open  & 1   & 4,316 \\
TOPDIAL \citep{wang2023target}             & LLM      & Act-topic pairs   & \textcolor{green}{\ding{51}} & \textcolor{red}{\ding{55}} & English    & 3    & 1   & 9,939 \\
\midrule
\rowcolor[gray]{0.92}
\textit{SOPDAIL} (ours)                           & LLM, Experts & Open Definition & \textcolor{green}{\ding{51}} & \textcolor{green}{\ding{51}} & Chinese & 32   & 53   & 3,114 \\
\bottomrule
\end{tabular}
}
\caption{
A comparison between our proposed \textit{SOPDAIL} and other relevant datasets, where PA indicates whether it includes proactive interaction, CT denotes controllability, and Lang denotes language.
}
\label{data-compare-table}
\end{table*}

\subsection{Dataset Curation Framework}
Inspired by recent studies \citep{wang2023target,sandler2024linguistic} that directly use LLMs to create high-quality dialogue datasets, we introduce a role-playing framework where LLMs simulate various agent roles to generate task-oriented dialogues. As shown in Figure \ref{fig: dataflow}, our framework follows a four-step curation process: task definition, SOP planning, dialogue path creation, and dialogue generation. The detailed steps are as follows.

\textbf{Step 1:} To satisfy the diversity of our dataset, we curate 53 unique tasks, including activities such as activating a bank card, scheduling appointments, and online shopping, across 32 domains, e.g., shopping, education, hospital, etc. Please refer to Appendix Table \ref{domain-task detail} for all tasks and domains.

\textbf{Step 2:} To generate intermediate SOP annotations for SOP prediction, as defined in \S\ref{taskdef}, we utilize zero-shot prompting to instruct LLMs in acting as \textbf{planners} to draft SOPs for the specified tasks. 

\textbf{Step 3:} We then prompt LLMs to generate multi-turn dialogue paths sampled from the annotated SOPs. To ensure the dialogues reflect proactive interactions and closely mimic real-world scenarios, we assign LLMs the role of \textbf{screenwriters}, instructing them to insert predefined proactive agent actions (e.g., offering help, persuading) and user states (e.g., asking question) into dialogue paths. 


\textbf{Step 4:} We assign LLMs as \textbf{scriptwriters} to draft dialogues for each agent action and user state, used to evaluate dialogue generation in \S\ref{taskdef}.

\textbf{Note:} Human annotators are involved in reviewing and refining the LLM-generated annotations in Steps 2-4, ensuring their accuracy and quality. Besides, a user simulator generates a unique user profile and updates the \textit{task definition} in step 3 and 4, enabling diverse dialogue paths and interactions. Detailed prompts are provided in Appendix \ref{all data create prompts}. 

\subsection{Human Annotation}
\label{human_evaluation}

\textbf{Annotator Selection.} We recruited seven annotators with relevant qualifications and expertise to ensure the quality of the annotation process. Before annotation, all participants are trained to gain a thorough understanding of the annotation guidelines.
The annotators worked independently but were allowed to provide feedback or reject any doubtful cases. On average, the annotation time was 10.3 minutes per sample, and annotators were compensated at a rate of \$8 per hour.  

\textbf{Annotation Process.} We develop an in-house web application as the annotation platform. Participants are required to read the guidelines, pass the pre-annotations, and then perform the actual annotations. Additionally, every instance was assigned to three annotators for cross-annotation validation with an inter-annotator agreement (IAA) score of 0.88, showing high consistency of annotation. Instances with an IAA below 0.95 were excluded. Please refer to Appendix \ref{annotator-details} for more details. 

\subsection{Dataset Analysis} 

\textbf{Comparison with Existing Datasets.}  
Table \ref{data-compare-table} provides a comparison of \textit{SOPDAIL} against other relevant datasets, highlighting the distinct advantages of our dataset. Notably, \textit{SOPDAIL} covers 53 tasks across 32 domains, offering a comprehensive and diverse evaluation framework for dialogue agents. Furthermore, this dataset is well-suited for in-context learning with LLMs~\citep{yu2023prompt}, as opposed to traditional fine-tuning approaches~\citep{kojima2022large}, thus enhancing the efficiency of task deployment. To the best of our knowledge, \textit{SOPDAIL} is the first Chinese benchmark for proactive and controllable dialogue. 
We anticipate this dataset will serve as a valuable resource for advancing research on controllable LLM-driven agents.

\textbf{Statistics.} 
Table \ref{statistics-table} summarizes the statistics of our \textit{SOPDAIL}. The quantity analysis shows that 74\% of the utterances are for controllability, aligning well with the objective of our work. To assess the quality of the dataset, we measured the accuracy of samples after annotation by human experts. Three annotators were invited to evaluate 300 randomly sampled cases, rating each as 1 if it adhered to the instructions and was semantically correct, and 0 otherwise. The results indicate a 0.98 accuracy score, underscoring the high quality of our dataset. Notably, even the raw results achieved a 0.91 accuracy rate, highlighting the superior performance of GPT-4o in this task.

\begin{table}[t]
\centering
\resizebox{\columnwidth}{!}{
\begin{tabular}{@{}lr@{}}
\toprule
\textbf{Quantity Statistics}\\
\midrule
Total \# Domains / Tasks / Goals & 32 / 53 / 70 \\
Total \# SOP vertices / Edges & 899 / 1,058 \\
Total \# Dialogues / Turns & 3,114 / 23,897 \\
Total \# Utterances / Tokens & 47,795 / 119,5736 \\
Avg. \# Turns / Utterances per dialogue & 7.67 / 15.34 \\
Avg. \# Words per utterances & 25.01 \\
\rowcolor[gray]{0.92}
Rate. \# Controllability / Proactivity & 0.74 / 0.26 \\
\midrule
\textbf{Quality Statistics} \\
\midrule
\multicolumn{2}{c}{\textit{Before Expert Correction}} \\
Avg. \# \textit{SOPDAIL} Sample Accuracy & 0.91 \\
\multicolumn{2}{c}{\textit{After Expert Correction}}\\
Avg. \# ED of dialogue paths / utterances & 0.41 / 0.34 \\
\rowcolor[gray]{0.92}
Avg. \# \textit{SOPDAIL} Sample Accuracy & 0.98 \\
\bottomrule
\end{tabular}
}
\caption{The statistics of our \textit{SOPDAIL}, where ED represents the edit distance used to evaluate the discrepancy between raw results and those after expert corrections. The high accuracy demonstrates the superior performance of LLMs and the high quality of our dataset.}
\label{statistics-table}
\end{table}

\section{Our Approach}
\begin{figure*}[t]
    \centering
    \includegraphics[width=1.0\linewidth]{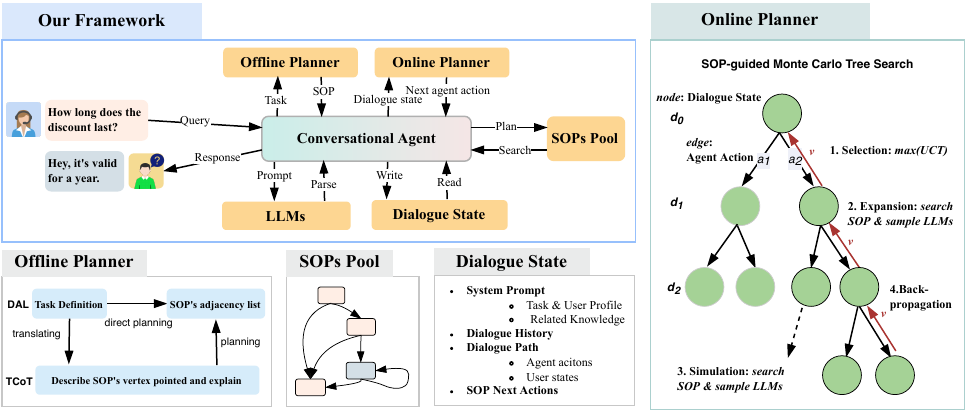}
    \vskip -0.5em
    \caption{Diagram of the ChatSOP framework. The system has five components: (1) \textbf{LLMs} Module for language models management; (2) \textbf{Dialogue State} for history management; (3) \textbf{SOPs Pool} for SOP storage; (4) the \textbf{Offline Planner}, which constructs the SOP graph before a conversation; and (5) the \textbf{Online Planner}, which generates dialogue in real-time using the SOP. \textbf{Workflow:} Before a conversation, the Offline Planner builds the SOP and stores it. During the conversation, the Online Planner retrieves the SOP to guide its MCTS-based response generation.}
    \label{fig: model}
\end{figure*}


In this section, we present the details of our proposed method, beginning with a multi-turn dialogue framework powered by LLMs. We then discuss the fine-tuning method for SOP prediction, followed by leveraging Monte Carlo Tree Search (MCTS) for dialogue generation.

\subsection{Framework Overview}
As illustrated in the left panel of Figure \ref{fig: model}, the planning-based dialogue agent (ChatSOP) consists of five components: 1) \textit{LLMs Module}: Responsible for managing and utilizing multiple LLMs to support various functionalities; 2) \textit{Dialogue State}: Handles the storage and update of task prompts and dialogue history.
3) \textit{SOPs Pool}: Contains pre-defined SOP vertices and predicted edges; 4) \textit{Offline Planner}: Constructs an adjacency list to assemble a complete SOP graph from the provided task definition; 5) \textit{Online Planner}: Generates the dialogue based on the predicted SOP graph.

Given a \textit{task definition} from the user, the agent initially retrieves relevant SOP nodes from the \textit{SOPs Pool} through an iterative search. The \textit{Offline Planner} is then employed to generate an adjacency list representing a complete SOP graph. Once the graph is constructed, the \textit{Dialogue State} module is activated to prepare task-specific prompts and manage dialogue history. Finally, the \textit{Online Planner} generates the dialogue using the SOP graph and the prepared prompts. It is important to note that both the Offline and Online Planners are powered by the \textit{LLMs Module}. Below, we present the implementation of the offline and online modules. 

\subsection{Offline Planner}

The objective of this module is to predict the adjacency List $\mathcal{M}$ based on user states and agent actions, thereby guiding user interactions with the agent. To achieve this, we propose three methods: Direct Adjacency List (DAL), Translation Chain-of-Thought (TCoT), and Supervised Fine-Tuning (SFT). Specifically, DAL employs direct prompting of LLMs to generate an adjacency list in JSON format. Here, an adjacency list represents the connectivity between two vertices, where a value of 1 indicates a connection and 0 indicates no connection. TCoT involves a two-step process: first, the LLMs are prompted to describe each vertex and its child vertices in natural language, including justifications for the relationships; then, the description is translated into an adjacency list in JSON format. Finally, SFT implements fine-tuning on various LLMs, such as Llama and Qwen, to iteratively generate the adjacency vertices for each vertex in the SOP graph. Please refer to Appendix \ref{prompt_details} for detailed prompt settings.


\subsection{Online Planner}
Following the prediction of the SOP graph, we now delve into dialogue generation driven by the online planner module. Specifically, the target is to predict the practical dialogue path, as shown in the right panel of Figure \ref{fig: model}, where nodes represent dialogue states $d_t$, including both agent actions $a_t$ and user states $s_t$. At each step, the agent is required to predict the next actions based on its working memory. However, direct use of exhaustive search over the entire space can lead to sub-optimal dialogue paths. Thus, in our work, we propose \textbf{S}OP-\textbf{g}uided \textbf{M}onte Carlo Tree Search (\textbf{SGM}) to construct the dialogue path, assuring to predict the optimal action through $N$ steps simulations.  

Given the initial dialogue state $d_0$, inspired by \citet{hao-etal-2023-reasoning}, we propose an iterative process to search for the optimal next action in 4 steps: node \textbf{selection}, node \textbf{expansion}, dialogue \textbf{simulation}, and \textbf{back-propagation}. After $n$ iterations, the optimal next action for $d_0$ is selected. This process continues until the predefined computational budget is reached (e.g., number of iterations), at which point the resultant trajectory can be extracted from the tree. The algorithm is provided in Appendix \ref{algorithm mcts}. Below, we provide the details of each stage.

\textbf{Selection.} Starting from the root node (i.e., the initial state $d_0$), a child node is selected at each level of the tree to determine the next state. This process continues until it reaches a leaf node.
To balance exploration (less-visited nodes) and exploitation (high-value nodes), we employ the Upper Confidence Bounds for Trees (UCT) algorithm \citep{kocsis2006bandit} for child node selection:

{\small
\begin{equation}
a_t^* = \arg\max_{a_t} \left[ Q(d_t, a_t) + w \sqrt{\frac{\ln N(d_t)}{N(c(d_t, a_t))}} \right]
\end{equation}
}
where \(N(d_t)\) denotes the number of times node \(d_t\) has been visited in previous iterations, \(c(d_t, a_t)\) is the child node of applying \(a_t\) in state \(d_t\) and $w$ denotes an empirical weight parameter (set to 1 in our experiments) to balance exploration and exploitation. The less a child node has been visited (i.e., the more uncertain the child node is), the higher the second term. 

The state-action value function \(Q(d_t, a_t)\) estimates the expected future reward associated with taking action \(a_t\) in state \(d_t\):

{\small
\begin{equation}
Q(d_t, a_t) = \lambda \cdot \left( \mathcal{L}(d_t, a_t) + \left(1 - \lambda\right) \cdot \mathrm{T}(d_t, a_t) \right)
\end{equation}
}
where $\mathcal{L}(d_t, a_t) (0 \leq \mathcal{L}(d_t,a_t) \leq 1) $ function represents the logical rationality of the current action, as assessed by the LLMs. It is computed as the mean of several binary (0 or 1) evaluations derived from prompt-based sampling by the LLMs. The $\mathrm{T}(d_t, a_t)$ function assigns discrete values to measure task completion: 0.3 for the termination state, 0.7 for the success state, and 0 for others. $\lambda$ is a hyperparameter, set to 0.5 in our experiments, that balances logical rationality and task completion.

 \textbf{Expansion.} After a leaf node (non-terminate) is selected, the agent samples $m$ possible dialogue states $d_t$ iteratively for expansion. Notably, we first utilize the local subgraph from SOP graph, then add the next two levels of child nodes connected to the current state node for further expansion. This setting ensures that the agent maintains a balance between constraints and proactivity. Finally, when the selected leaf node is already a terminal node (either a dialogue end node or the maximum search depth has been reached), we will skip the expansion phase and proceed to back-propagation.

\textbf{Simulation.} To estimate the reward generated by future dialogue, we simulate the future dialogue for each expanded state node. To improve efficiency, we follow a process similar to the expansion phase mentioned above, that is, we only simulate downward for candidate dialogue policies that are sampled from LLM  and guided by the SOP.

\textbf{Backpropagation.} At the final step, once a terminal state is reached, the 
$Q$ values are updated along the entire dialogue path. The algorithm terminates when the predetermined total number of iterations is completed. Finally, within the constructed dialogue tree, the child node with the highest $Q$ value of the current node is selected to guide the next turn in the conversation. The details are provided in Algorithm~\ref{algorithm mcts} in Appendix~\ref{sec-algorithm}.

\begin{table}[t]
\centering
\scalebox{0.95}{
\begin{tabular}{l|ccc}
\toprule
\textbf{Task} & \textbf{Train} & \textbf{Valid} & \textbf{Test} \\
\midrule
SOP Prediction & 31 & 5 & 17 \\
\midrule
Dialogue Generation & 1,859 & 324 & 931 \\
\bottomrule
\end{tabular}}
\caption{
   \label{datasplit-table} 
   \textit{SOPDAIL} dataset statistics for training, validation, and test splits across different domain tasks. SOP Prediction utilizing 5-fold cross-validation.}
\end{table}

\section{Experiments} 
We conducted extensive experiments to validate our curated dataset and planning-based dialogue agent with both automatic and human evaluations. 

\subsection{Datasets}
We split \textit{SOPDAIL} dataset into training, validation, and test sets, as shown in Table \ref{datasplit-table}. To evaluate the methods on unseen tasks, we split the data at the task level rather than the dialogue level, thereby avoiding any task overlap between the three sets. Additionally, we employed 5-fold cross-validation for SOP prediction to ensure result validity.

\subsection{Experimental Setting}

\textbf{Baseline Setting} For SOP prediction, we compare our method (TCoT and SFT) against the baselines DAL. For dialogue generation, we compare our method (SGM) with CoT and CoT+SOP. The LLMs include GPT-3.5-turbo/4o, Qwen1.5-14b/72b-chat, and Llama3-8b/70b-chat, covering both open and closed models across different sizes.

\textbf{Paramerters} We use the GPT models\footnote{GPT: \url{https://platform.openai.com/docs/models}} through the provided API, while for open-source models, we directly load the pre-trained versions from HuggingFace models\footnote{HuggingFace: \url{https://huggingface.co/models}}, experiments on 4 Nvidia A800 GPUs. The inference is performed with a temperature and top-p setting of 0.1. For SFT, we fine-tuned all parameters using 5 epochs, 50 warm steps, 1e-6 learning rate, and 128 batch size. Hyperparameters are in Appendix \ref{sec-exper-details}

\textbf{Evaluation Metrics} 
We evaluated SOP prediction in terms of graph structure and dialogue usability. For graph structure, we calculated the graph edit distance (GED) and its operation ratio (GEDR) \cite{hagberg2008exploring} needed to match the predicted SOP with the ground truth. For dialogue usability, we measured the precision, recall, and F1 score of the dialogue paths, defining true positives as paths in the predicted SOP that exactly match the ground-truth paths.
For dialogue generation, we measured the accuracy of single-turn dialogue actions (Acc T), controllable actions (Acc C), and proactive actions (Acc P) to evaluate the dialogue's controllability and proactivity. The accuracy of the entire dialogue (Acc D) was calculated by dividing the number of correct dialogues (i.e., each turn is correct) by the total number of dialogues. 

\subsection{Main Results}


\begin{table}[t]
\centering
\scalebox{0.67}{
\begin{tabular}{llccccc}
\toprule
\textbf{Model} & \textbf{Method} & \textbf{Pre}$\uparrow$ & \textbf{Rec}$\uparrow$ & \textbf{F1}$\uparrow$ & \textbf{GED}$\downarrow$ & \textbf{GEDR}$\downarrow$ \\
\midrule
\multirow{2}{*}{GPT-4o} & DAL & \cellgrey{\textbf{78.11}} & 66.51 & \cellgrey{\textbf{71.85}} & \cellgrey{\textbf{2.01}} & \cellgrey{\textbf{5.40}} \\ 
                            & TCoT & 69.20 & \cellgrey{\textbf{73.34}} & 71.22 & 4.70 & 12.46 \\ 
\midrule
\multirow{2}{*}{GPT-3.5} & DAL & 41.39 & 21.94 & 28.68 & 7.61 & 22.48 \\
                       & TCoT & \cellgrey{\textbf{50.95}} & \cellgrey{\textbf{50.03}} & \cellgrey{\textbf{50.48}} & \cellgrey{\textbf{6.91}} & \cellgrey{\textbf{18.28}} \\
\midrule
\multirow{3}{*}{Qwen1.5-14b} & DAL & 42.84 & 37.81 & 40.17 & 10.23 & 25.77 \\
                       & TCoT & 38.79 & 42.69 & 40.65 & 11.36 & 26.91 \\
                       & SFT  & \cellgrey{\textbf{68.58}} & \cellgrey{\textbf{62.73}} & \cellgrey{\textbf{65.52}} & \cellgrey{\textbf{3.94}} & \cellgrey{\textbf{10.30}} \\
\midrule
\multirow{3}{*}{Qwen1.5-72b} & DAL & 46.74 & 31.57 & 37.68 & 10.75 & 28.32 \\
                       & TCoT & 48.29 & 51.94 & 50.04 & 7.38 & 18.72 \\
                       & SFT  & \cellgrey{\textbf{80.25}} & \cellgrey{\textbf{74.01}} & \cellgrey{\textbf{77.00}} & \cellgrey{\textbf{2.86}} & \cellgrey{\textbf{7.12}} \\
\midrule
\multirow{3}{*}{Llama3-8b} & DAL & 35.19 & 28.04 & 31.21 & 8.31 & 24.05 \\
                       & TCoT & 44.50 & 40.35 & 42.32 & 10.92 & 30.10 \\
                       & SFT  & \cellgrey{\textbf{72.19}} & \cellgrey{\textbf{68.33}} & \cellgrey{\textbf{70.21}} & \cellgrey{\textbf{3.54}} & \cellgrey{\textbf{9.33}} \\
\midrule
\multirow{3}{*}{Llama3-70b} & DAL & 64.14 & 56.05 & 59.82 & 3.85 & 10.72 \\
                       & TCoT & 60.81 & 67.34 & 63.91 & 5.43 & 14.38 \\
                       & SFT  & \cellgrey{\textbf{74.10}} & \cellgrey{\textbf{73.03}} & \cellgrey{\textbf{73.56}} & \cellgrey{\textbf{2.81}} & \cellgrey{\textbf{7.27}} \\
\bottomrule
\end{tabular}
}
\caption{Results for SOP prediction, where Pre and Rec are precision and recall, while GED and GEDR are the graph edit distance and its editing ratio, showing that SFT significantly outperforms the baselines.}
\label{task1-table}
\end{table}

\begin{figure*}[t]
    \centering
    \includegraphics[width=1\linewidth]{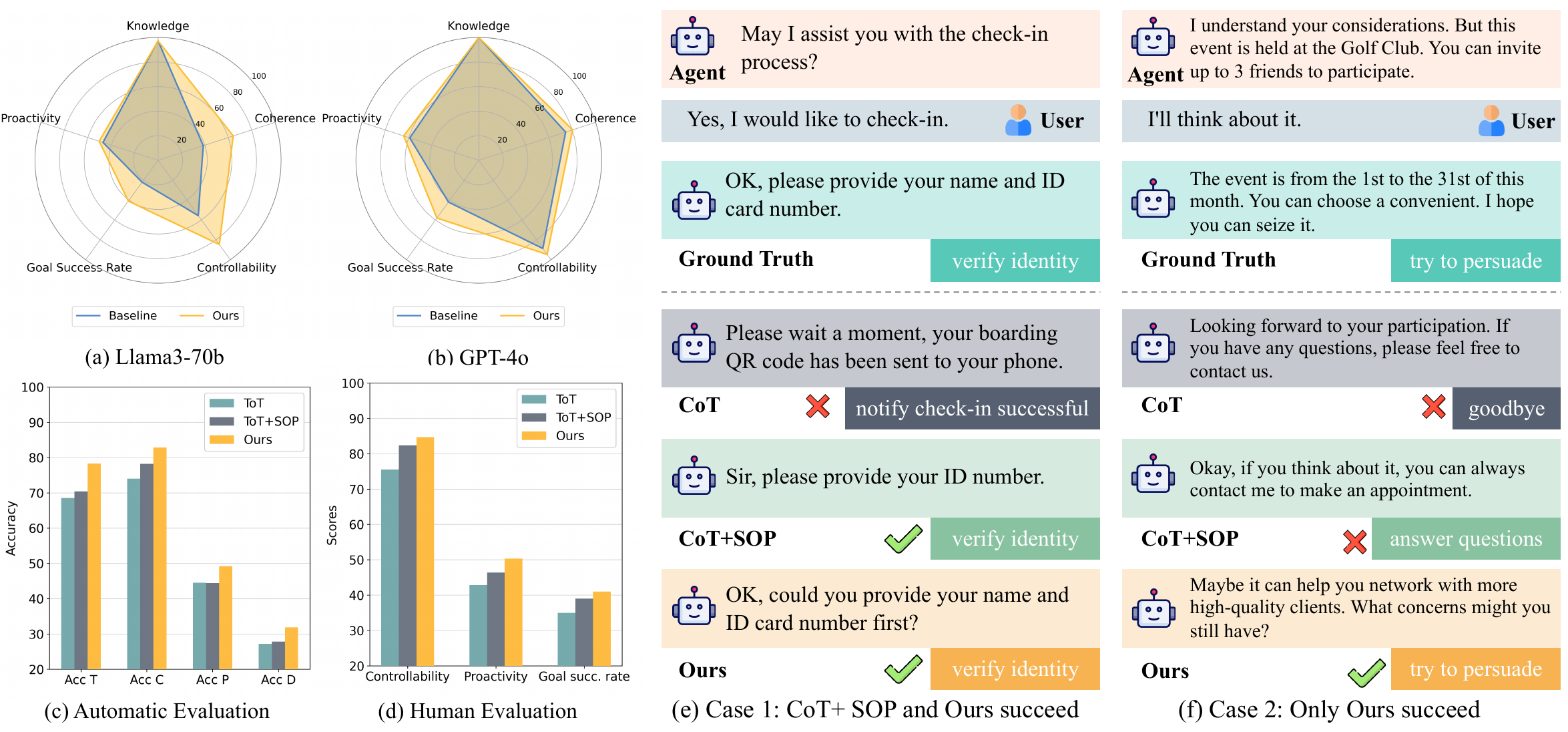}
    \vskip -0.5em
    \caption{More analysis results: (a)-(b) presenting human evaluation results for Llama3-70b and GPT-4o models based on the five proposed metrics; (c)-(d) comparing ToT with our method (SGM) through both automatic and human evaluations, demonstrating the superiority of our approach; and (e)-(f) providing two case studies to further illustrate the advantages of our method.}
    \label{case_fig}
    \vskip -0.5em
\end{figure*}
\textbf{SOP Prediction} Table \ref{task1-table} presents the results of our methods (TCoT and SFT) compared to DAL. The TCoT outperforms the baseline DAL on most metrics, particularly showing significant improvements in recall and F1 scores. This indicates that TCoT's predicted SOP has a stronger advantage in dialogue usability.
Similarly, our SFT method also shows significant improvements over all metrics and models. Notably, the fine-tuned Qwen1.5-72b achieves the highest score, even surpassing GPT-4o (\textit{77.00} vs \textit{71.85} in F1 score), underscoring the effectiveness of our proposed dataset.

\textbf{Dialogue Generation} For dialogue action generation, as shown in Table \ref{task2-table}, our method (SGM) substantially outperforms the baseline methods across all metrics and models. GPT-3.5 based agent achieves a score of \textit{76.52} compared to \textit{48.57} by the baseline. Additionally, larger models demonstrate better performance, with GPT-4o achieving the highest scores, particularly excelling in controllable action generation with a score of \textit{91.19}. These findings highlight the effectiveness of our proposed method in tackling such a challenging multi-scenario dataset and generating dialogues.

\begin{table}[t]
\vskip 0.6em 
\centering
 \scalebox{0.67}{
\begin{tabular}{llcccc}
\toprule
\textbf{Model} & \textbf{Method} & \textbf{Acc T} $\uparrow$ & \textbf{Acc C} $\uparrow$ & \textbf{Acc P} $\uparrow$ & \textbf{Acc D} $\uparrow$ \\
\midrule
\multirow{3}{*}{GPT-4o} & CoT        & 69.64 & 74.34 & 51.30 & 22.72 \\ 
                        & CoT+SOP    & 82.09 & 86.73 & 57.14 & 43.39 \\ 
                        & SGM (our)   & \cellgrey{\textbf{86.37}} & \cellgrey{\textbf{91.19}} & \cellgrey{\textbf{60.42}} & \cellgrey{\textbf{46.29}} \\ 
\midrule
\multirow{3}{*}{GPT-3.5} & CoT   & 48.57 & 55.75 & 18.52 & 4.83  \\ 
                        & CoT+SOP & 63.24 & 69.49 & 22.22 & 28.57 \\ 
                        & SGM (our) & \cellgrey{\textbf{76.52}} & \cellgrey{\textbf{80.42}} & \cellgrey{\textbf{33.10}} & \cellgrey{\textbf{38.45}} \\ 
\midrule
\multirow{3}{*}{Qwen1.5-14b} & CoT    & 50.47 & 55.75 & 25.93 & 2.15  \\ 
                                & CoT+SOP  & 55.64 & 61.57 & 25.67 & 2.69  \\ 
                                & SGM (our) & \cellgrey{\textbf{61.76}} & \cellgrey{\textbf{67.11}} & \cellgrey{\textbf{28.75}} & \cellgrey{\textbf{3.33}} \\
\midrule
\multirow{3}{*}{Llama3-8b} & CoT & 31.79 & 36.73 & 11.11 & 0.32 \\ 
                        & CoT+SOP & 38.72 & 43.28 & 17.14 & 2.26\\ 
                        & SGM (our) & \cellgrey{\textbf{46.85}} & \cellgrey{\textbf{56.30}} & \cellgrey{\textbf{22.31}} & \cellgrey{\textbf{3.76}} \\ 
\midrule
\multirow{3}{*}{Qwen1.5-72b} & CoT & 68.57 & 73.89 & 46.30 & 12.57 \\ 
                            & CoT+SOP   & 74.25 & 79.65 & 45.24 & 30.72 \\ 
                            & SGM (our)  & \cellgrey{\textbf{77.83}} & \cellgrey{\textbf{83.54}} & \cellgrey{\textbf{47.32}} & \cellgrey{\textbf{29.32}} \\ 
\midrule
\multirow{3}{*}{Llama3-70b} & CoT & 65.43 & 72.52 & 44.22 & 11.71 \\ 
                        & CoT+SOP  & 52.24 & 54.42 & 40.48 & 12.24 \\ 
                        & SGM (our) & \cellgrey{\textbf{78.35}} & \cellgrey{\textbf{82.86}} & \cellgrey{\textbf{49.18}} & \cellgrey{\textbf{31.87}} \\ 
\bottomrule
\end{tabular}
}
\caption{Automatic evaluation results of dialogue generation. T, C, P, and D denote turn, controllable, proactive, and dialogue, respectively.}
\label{task2-table}
\vskip -0.8em 
\end{table}


\subsection{Human Evaluations}

To further assess the quality of generated utterances, we conducted a human evaluation to measure the proportion of accurate control actions, proactive actions, and knowledge accuracy in each single-turn on 100 dialogues sampled from compare models. Besides, from a broader dialogue perspective, we further proposed two additional metrics: goal success rate, defined as correct actions, correct knowledge and goal achieved, and logical coherence score, defined as logic correct and consistent to history, both scored on a scale from 0 to 1. Detailed definitions are listed in Appendix \ref{manually_evaluate}. The evaluation was conducted by the same annotators previously described in $\S$\ref{human_evaluation}.

As illustrated in Figure \ref{case_fig}(a)-(b), we compare the performance of our method with the baseline across two models. For Llama3-70b, our method demonstrates significantly superior performance, with particularly notable improvements in controllability, goal success rate, and coherence. For GPT-4o, while the baseline model already achieves strong results, our method can still enhance scores across multiple dimensions, underscoring its effectiveness. Notably, in the knowledge dimension, all methods achieve consistently high scores, indicating that the models possess sufficient knowledge to support task-oriented dialogues effectively.

\begin{figure}[t]
    \centering
    \includegraphics[width=1\linewidth]{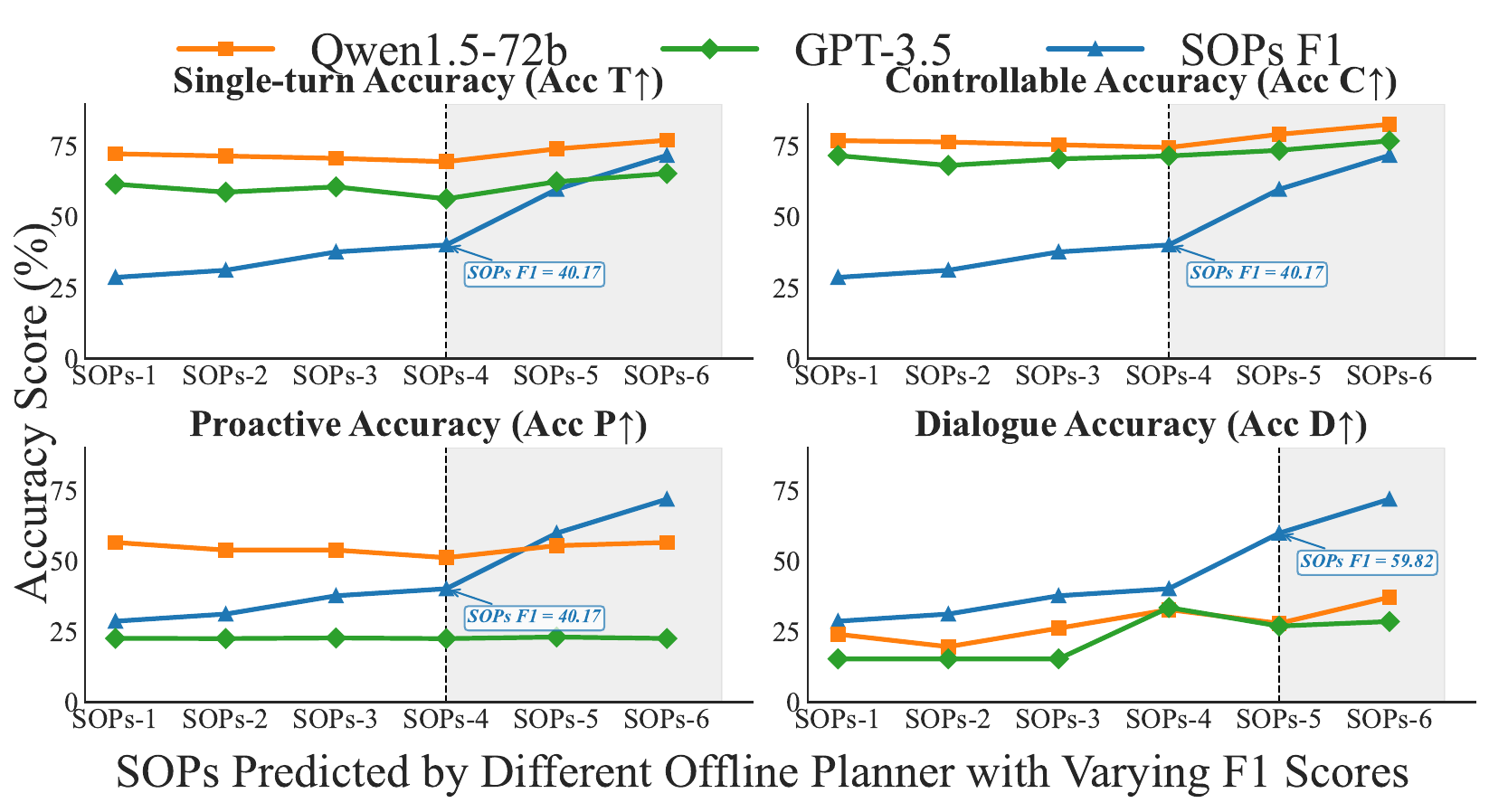}
    \vskip -0.5em
    \caption{
    Impact of SOP's accuracy (F1) on dialogue evaluation metrics for our SGM with different LLMs. The gray area indicates that the SOP offline planner’s improvement can enhance the performance of SGM.
    }
    \label{sop_fig}
    \vskip -0.8em
\end{figure}
\subsection{Analysis Experiments}


\textbf{Effect of Offline Planner} To better understand how offline planners affect SGM, we selected SOPs with varying F1 scores from the results of different online planners as input for SGM’s dialogue prediction. 
Figure \ref{sop_fig} shows the correlation between four dialogue evaluation metrics and SOP-F1.
The statistical results reveal strong positive Spearman correlations between SOP-F1 and overall dialogue accuracy (Acc D): 0.88 for Qwen and 0.75 for GPT. 
More specifically, it can be observed from see shaded area in Figure \ref{sop_fig} that the impact of SOPs is negligible when their F1 scores are below 40.17 or 59.82, but progressively increases above these thresholds. 
This indicates that higher SOP accuracy corresponds to better dialogue performance.

\textbf{Effect of Online Planner} To evaluate the performance of different tree search algorithms for online planner, we implemented Tree-of-Thought (ToT) for comparison with Monte Carlo Tree Search (MCTS).
Figures \ref{case_fig}(c)-(d) show that our SGM yields superior results in automatic and human evaluations. Moreover, incorporating SOP with ToT enhances performance, emphasizing the consistent benefits of SOP integration. 

\textbf{Cost-Effectiveness Analysis} 
We analyze the cost-effectiveness of different online planning algorithms with the same SOP offline planner, by measuring average \textit{token}, \textit{time}, and \textit{monetary} costs per conversational turn on LLaMA3-70B-Chat.
Compared to CoT, SGM improves goal success rate by 19\% with a cost increase from \$0.055 to \$0.456 per turn (roughly 8×).
Compared to ToT, SGM yields a slightly higher success rate (41\% vs. 39\%) while being notably more efficient—saving 8.64 seconds and \$0.224 per turn. These results suggest that SGM achieves a better trade-off between performance and efficiency.

\begin{figure}[t] 
    \centering
    \includegraphics[width=0.8\linewidth]{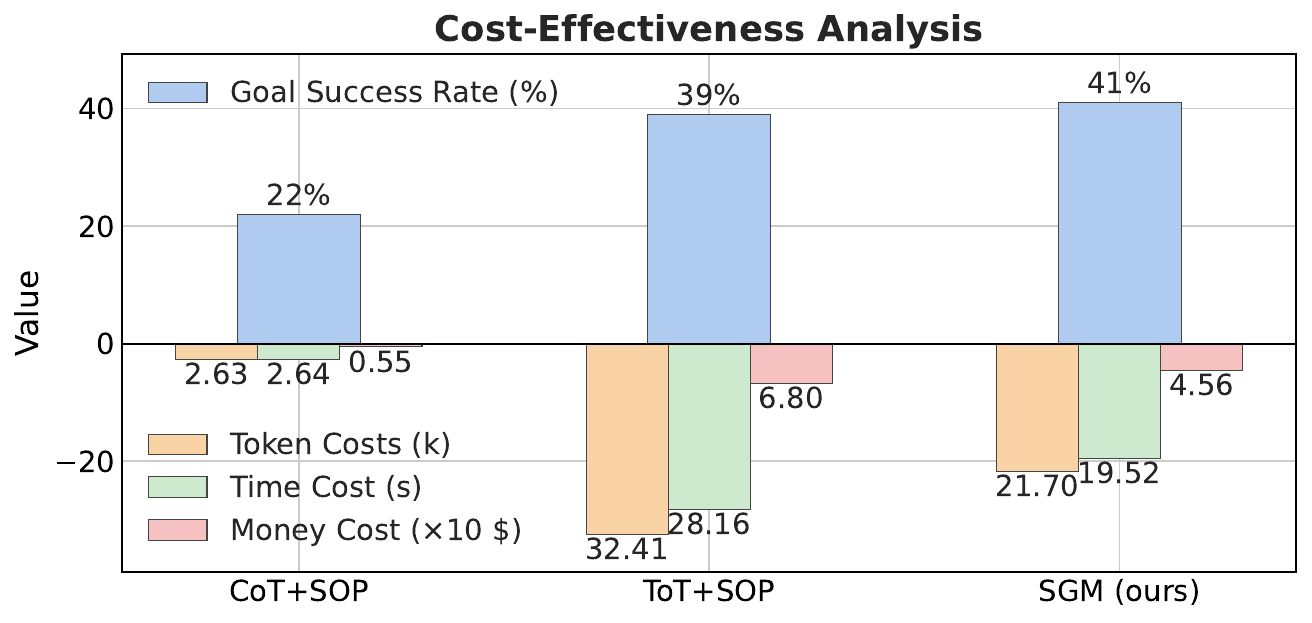} 
    \vskip -0.5em 
    \caption{Comparison of cost-effectiveness among different online planners per conversational turn. \textit{Money Costs} are scaled ×10 for better visibility.}
    \label{fig:cost} 
\end{figure}

\textbf{Comparison with RL-based Dialogue Policy}
We conducted a comparative experiment with the state-of-the-art RL-based dialogue policy method PDPD \citep{he-etal-2024-planning}, which also utilizes dual planners: an LM-based planner and an MCTS-based planner activated under uncertainty. For fair comparison, we adapted PDPD to an unsupervised setting consistent with our SGM framework, with aligned hyperparameters. The results from Table \ref{pdpd-table} show that our SGM method achieves consistently higher accuracy across all dialogue policy metrics, especially in controllable policy selection.

\subsection{Case Study}
Furthermore, Figures \ref{case_fig}(e)-(f) provide case studies comparing the dialogue actions selected and responses generated by different methods in the same contexts. In the first case, verifying identity is a prerequisite before checking in. After applying SOP, both CoT and ours select actions aligned with SOP guidelines. However, in scenarios where the optimal action is absent from the SOP, such as proactive persuasion in the second case, CoT+SOP fails to continue the persuasion attempt. In contrast, our method can leverage simulation and deeper dialogue path exploration to select a more goal-oriented action, resulting in a more effective persuasion strategy.

\begin{table}[t]
\centering
 \scalebox{0.67}{
\begin{tabular}{llcccc}
\toprule
\textbf{Model} & \textbf{Method} & \textbf{Acc T} $\uparrow$ & \textbf{Acc C} $\uparrow$ & \textbf{Acc P} $\uparrow$ & \textbf{Acc D} $\uparrow$ \\
\midrule
\multirow{2}{*}{GPT-4o} 
                        & DPDP       & 72.29 & 72.81 & 56.51 & 24.50 \\
                        & SGM (our)  & \cellgrey{\textbf{86.37}} & \cellgrey{\textbf{91.19}} & \cellgrey{\textbf{60.42}} & \cellgrey{\textbf{46.29}} \\ 
\midrule
\multirow{2}{*}{Qwen1.5-72b} 
                            & DPDP & 69.92 & 72.79 & 47.27 & 16.05 \\
                            & SGM (our)  & \cellgrey{\textbf{77.83}} & \cellgrey{\textbf{83.54}} & \cellgrey{\textbf{47.32}} & \cellgrey{\textbf{29.32}} \\ 

\bottomrule
\end{tabular}
}
\caption{Automatic evaluation results of PDPD and our SGM. T, C, P, and D denote turn, controllable, proactive, and dialogue, respectively.}
\label{pdpd-table}
\vskip -0.8em 
\end{table}

\section{Conclusion}
In this paper, we have presented a planning-based framework, a high-quality benchmark dataset, and an unsupervised algorithm that encodes SOP constraints into Monte Carlo Tree Search for controllable LLM dialogue agents. Based on LLMs without additional training, our approach offers a better controllable and scalable solution for enterprise-level dialogue systems.

\section*{Limitations}
\textbf{Hallucinations}  Our approach is based on the context learning of LLMs, such as ChatGPT and GPT-4. As LLMs may produce outputs containing hallucinations \citep{bang2023multitask}, our system might provide information beyond the task definition. We intend to enhance the veracity of responses through post-processing steps, such as training a dedicated safety model and incorporating checks and revisions into the post-processing phase.
\\
\textbf{Runtime} One significant limitation of our method is the runtime. The more exhaustive the tree search is (e.g., increasing n or k), the more likely the algorithm is to find the optimal dialogue policy. However, this comes at the cost of longer simulation times, which may impact the overall user experience. We believe that parallelizing the tree search or reusing portions of the simulated subtrees could help to speed up the runtime. We anticipate that with the advancement of LLMs research, the speed of inference will continue to improve. 

\section*{Ethics Statement}
Given the independent behavior of agents in goal-oriented dialogue, it's imperative to scrutinize ethical implications. Our approach does not force the agent to achieve a specified goal, nor does it force the user to accept the agent's request. Instead, our work highlights the criticality of directing agents to adhere to human-defined limitations. While our measures are potent, we advocate for the stringent regulation of goal signals, particularly when implementing goal-oriented dialogue systems in specialized fields. Currently, the targeting process must uphold factual accuracy, respect user privacy norms, and comply with societal laws.

\section*{Acknowledgments}
The present research was supported by the National Key Research and Development Program of China (Grant No. 2024YFE0203000). 
We would like to thank all the anonymous reviewers for their constructive and insightful feedback, which has greatly helped to improve the quality of this paper.

\bibliography{anthology,custom}

\input{appendix}

\end{document}

%% file: appendix.tex
\appendix
\clearpage
\section{Appendix}
\subsection{Task Definition}
\label{apdx: task_define_example}
Task Definition for dialogue should include the definitions of fields $task\_profile$, $user\_profile$, $agent\_action$, and $user\_state$. The definitions of these fields are as follows:
\begin{itemize}
\item $task\_profile$: Task information for providing business content, accomplishment goals, and relevant background knowledge to Agents. The example is as follows:

\begin{figure}[h] 
    \centering
    \includegraphics[width=0.45\textwidth]{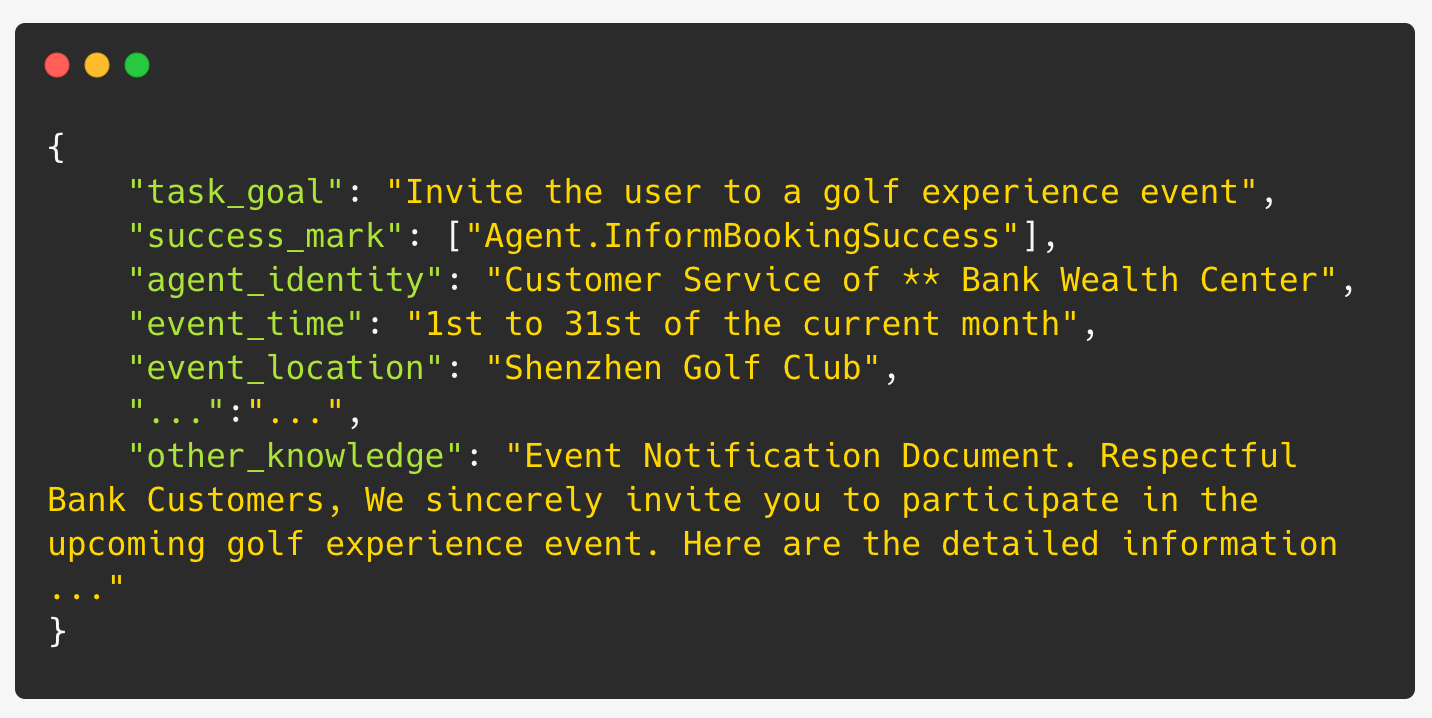} 
    \caption{Task profile prompt setting.}
    \label{fig:json_1} 
\end{figure}

\item $user\_profile$: The information about user business and personal information held by agents generally comes from the company's user management system. This information is used for identity verification or providing personalized services in conversations. The following is a example:

\begin{figure}[h] 
    \centering
    \includegraphics[width=0.45\textwidth]{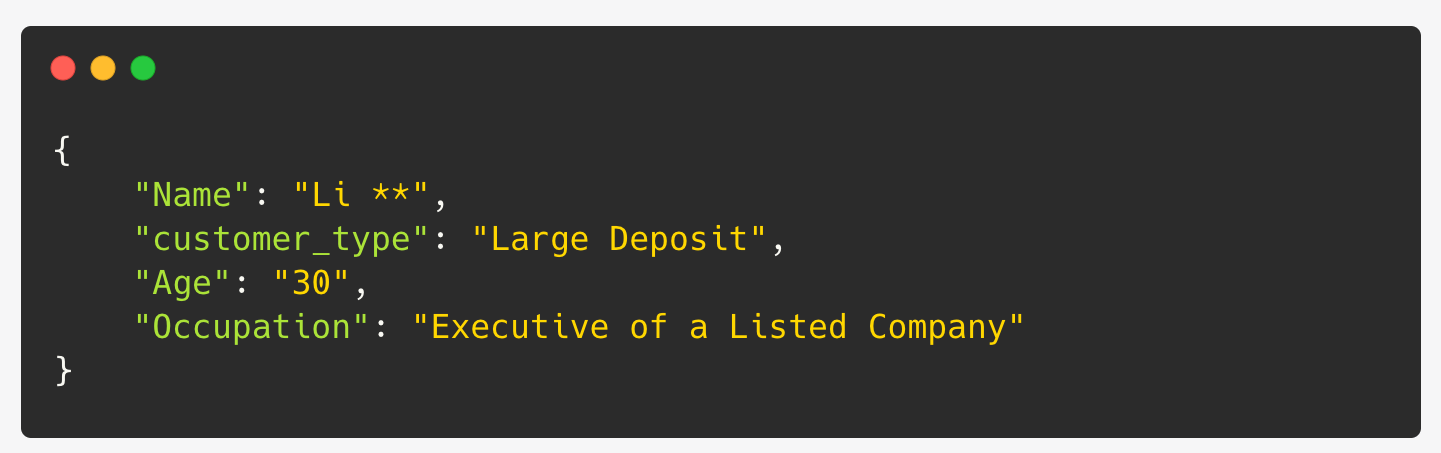} 
    \caption{User profile prompt setting.}
    \label{fig:json_2} 
\end{figure}


\item $agent\_action$: Summary of key actions to be carried out during the process of agent dialogue, intended to guide and constrain the content of the agent's dialogue in accordance with business regulations. Typically corresponds to the node names in the SOP or the strategy names of proactive dialogue. Here are some examples:

\begin{figure}[h] 
    \centering
    \includegraphics[width=0.45\textwidth]{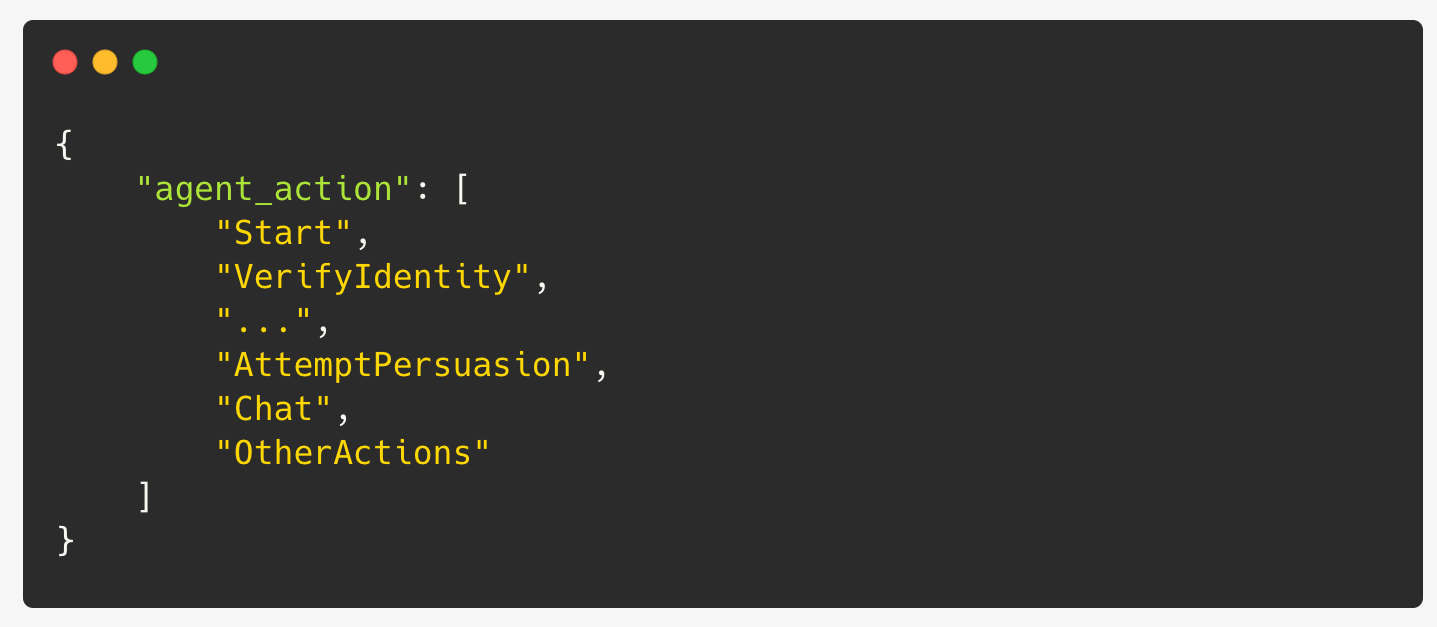} 
    \caption{Agent actions setting.}
    \label{fig:json_3} 
\end{figure}

\item $user\_state$: The status of the task summary from the user's final response combined with the preceding dialogue, serves as a prompt for the agent to select the optimal next action. This typically corresponds to the user node in the SOP or the proactive dialogue state of the user. The following are examples:

\begin{figure}[h] 
    \centering
    \includegraphics[width=0.45\textwidth]{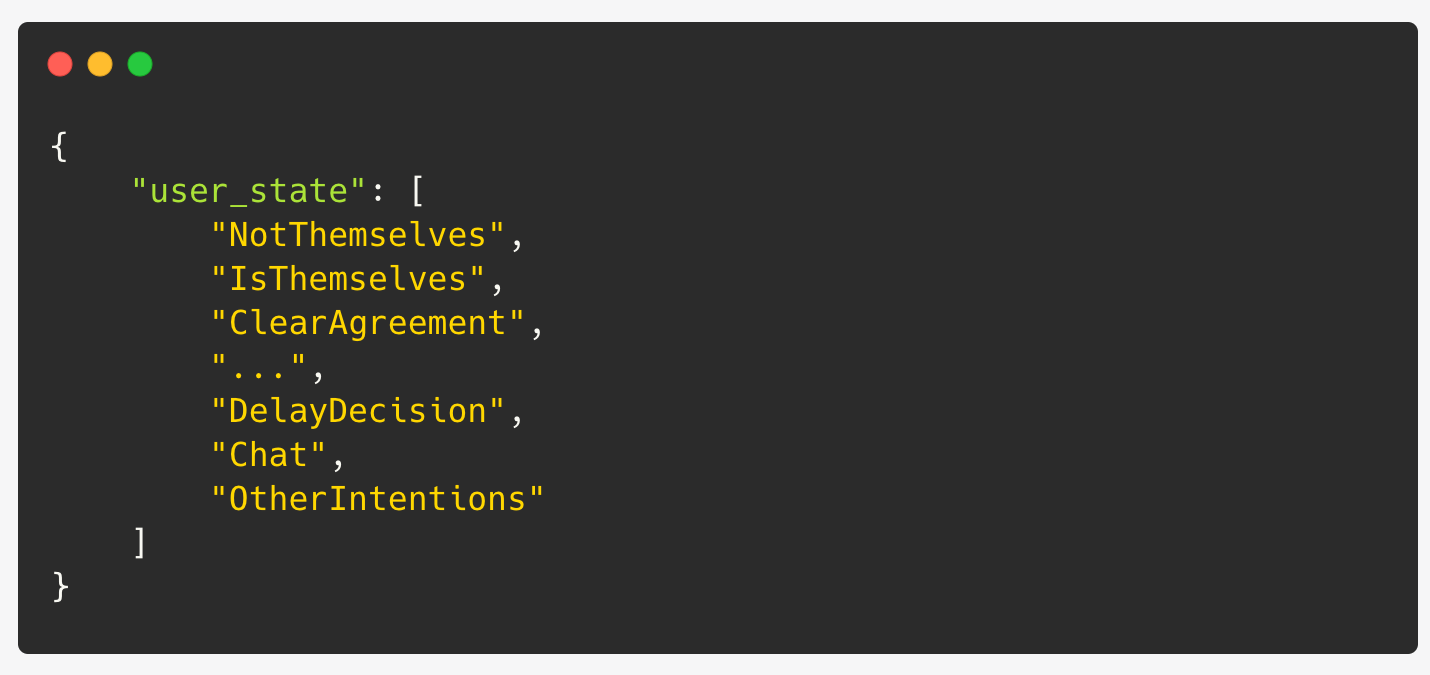} 
    \caption{User States setting.}
    \label{fig:json_4} 
\end{figure}

\end{itemize}

\begin{figure}[!t]
    \centering
    \includegraphics[width=1\linewidth]{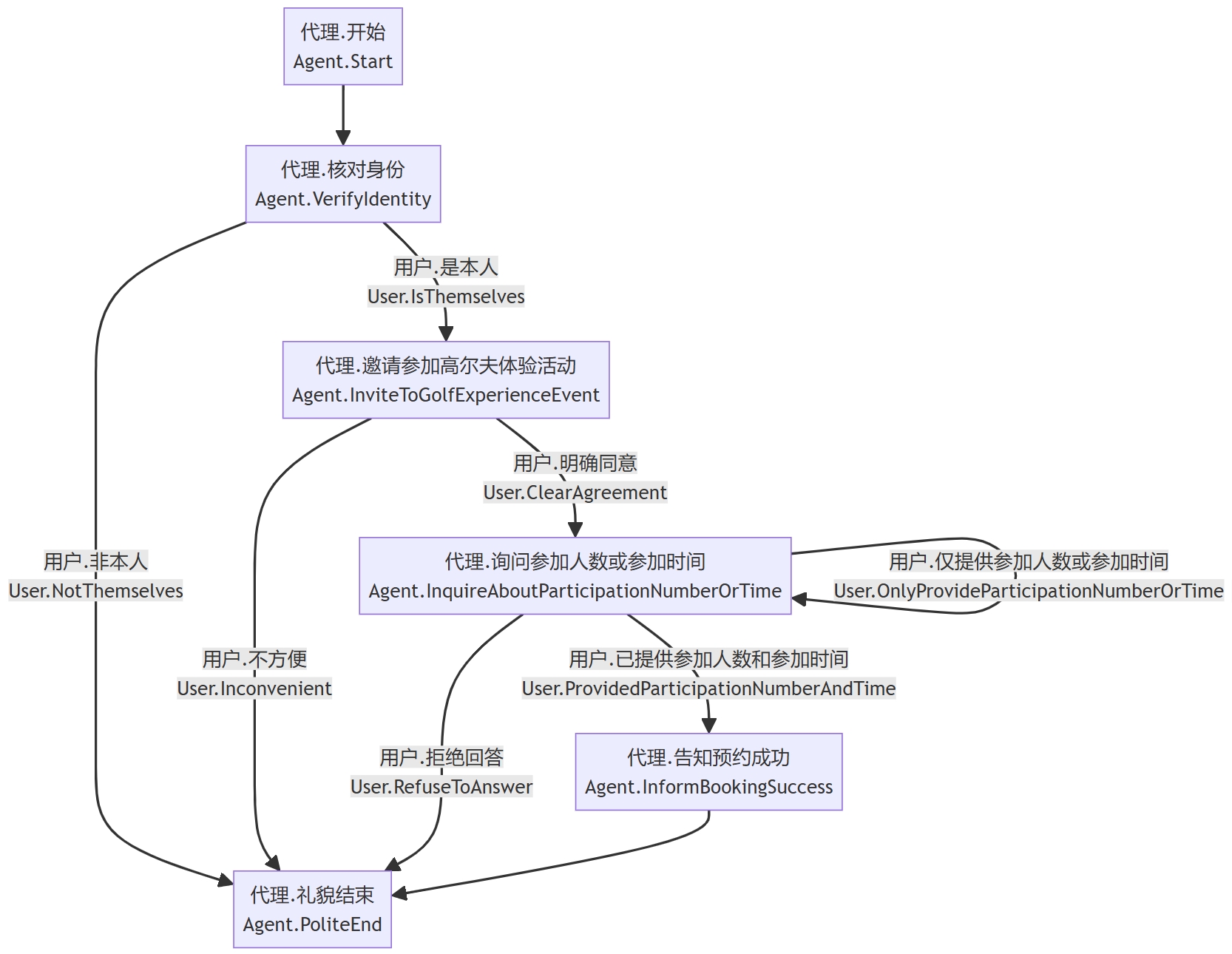}
    \vskip -0.5em
    \caption{An example of SOP definition.}
    \label{fig: task_define}
\end{figure}

\subsection{SOP Definition}
\begin{figure}[h] 
    \centering
    \includegraphics[width=0.45\textwidth]{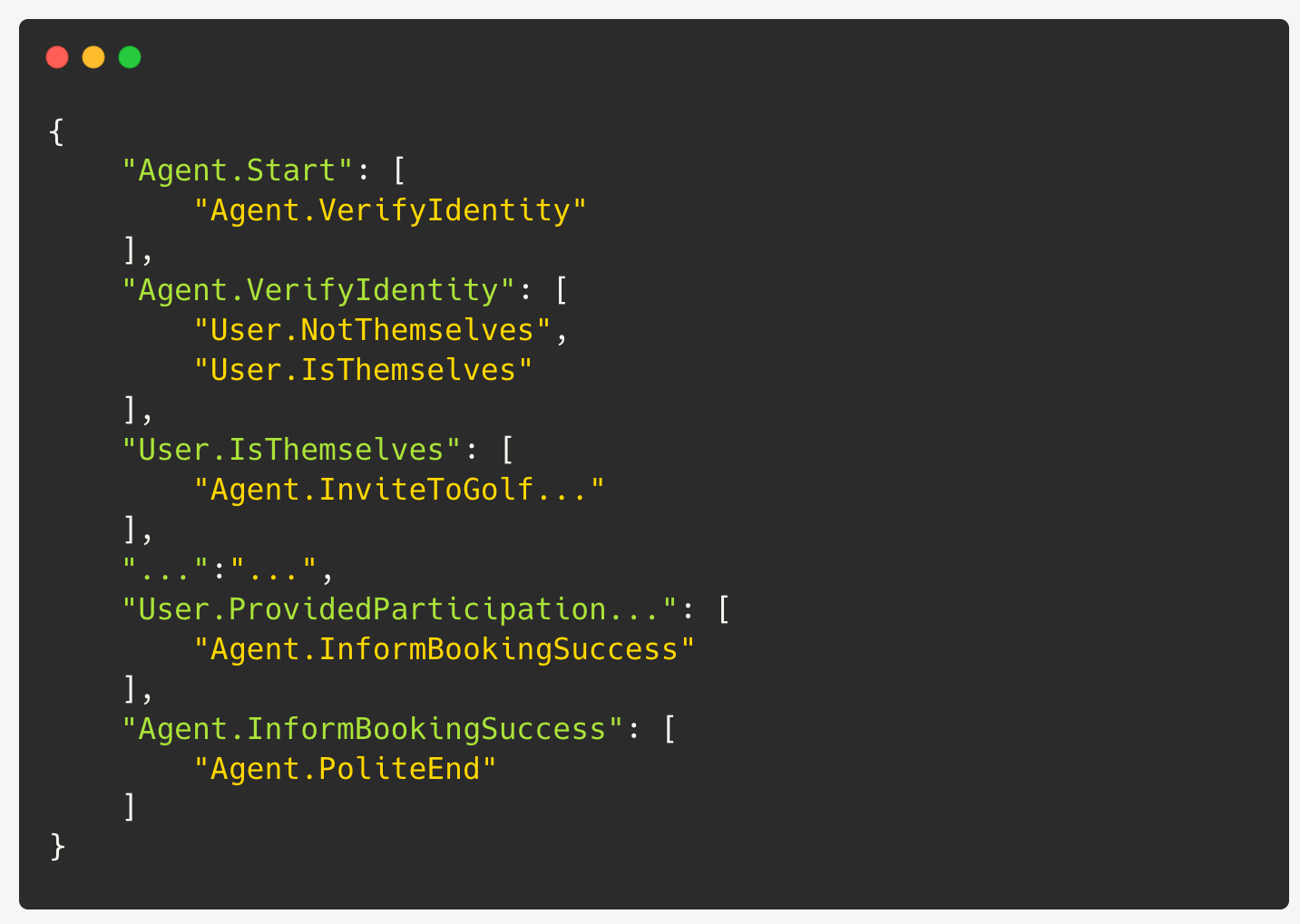} 
    \caption{The adjacency list representation of the SOP.}
    \label{fig:json_5} 
\end{figure}

Standardized operating procedures (SOP) is a directed graph where vertexes are $agent\_action$ and $user\_state$, and the edges
indicate the connections between these nodes. The SOP is established by business experts to standardize the essential business processing steps. Omitting or altering these steps can result in a violation or error. For instance, in the ``Activate Credit Card Invitation'' task, the agent must first ``verify the user's identity'' and confirm that the user is a ``cardholder'' before introducing the ``activation activity.'' Conversely, steps that do not impact the business process, such as the user ``asking questions'' and the agent ``resolving doubts,'' should not be included in the SOP. When actions are included in the SOP, they are referred to as controllable actions; otherwise, they are termed proactive actions. Figure \ref{fig: task_define} shows an example of SOP graph. The adjacency list representation of the SOP is shown in Figure \ref{fig:json_5}.


\subsection{Dataset Curation Details}
\label{dataset-details}

\begin{figure*}[t]
    \centering
    \includegraphics[width=0.95\linewidth]{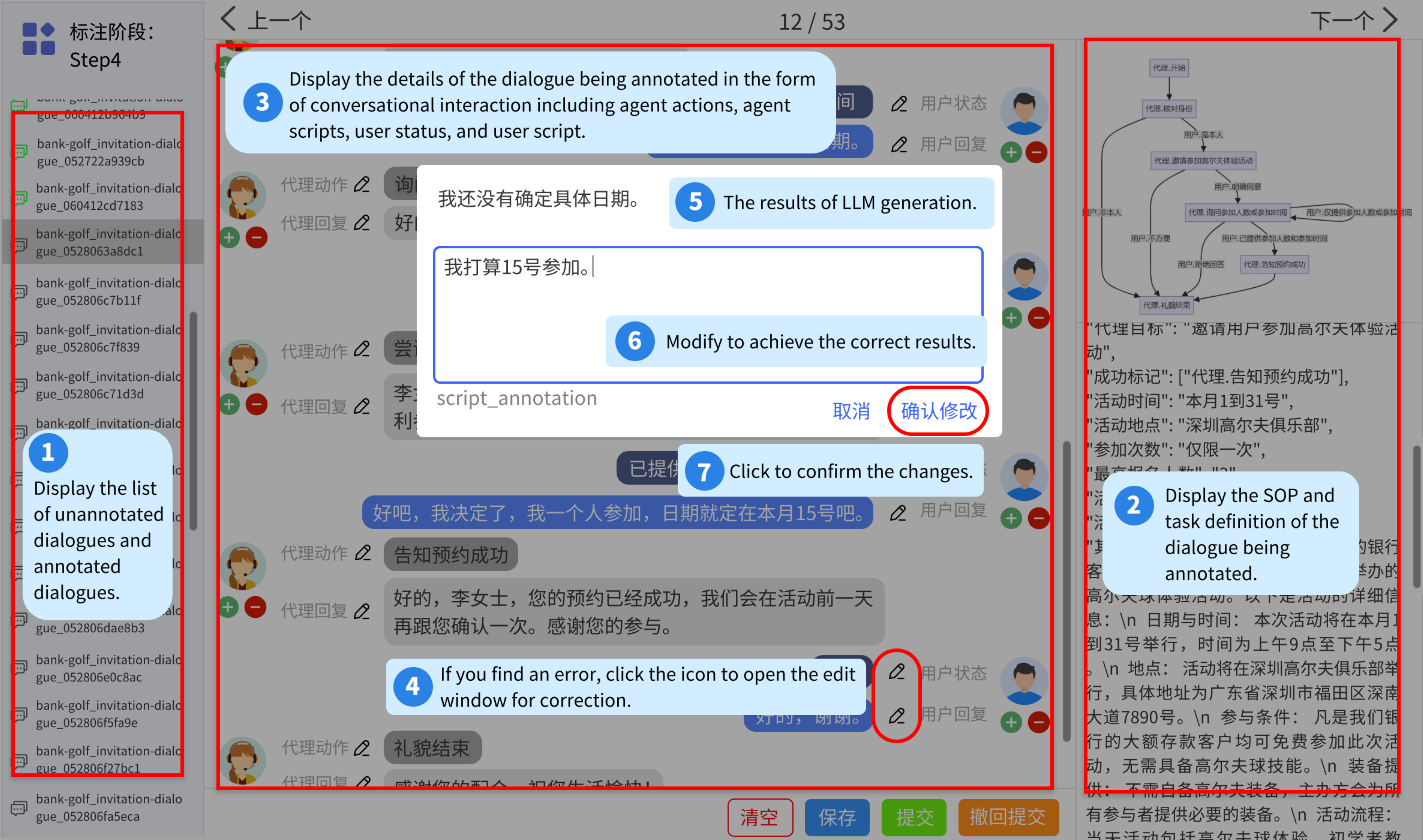}
    \vskip -0.5em
    \caption{Screenshot of the annotation tool.}
    \label{annotation_tool_fig}
\end{figure*}

\subsubsection{Prompts of Role-Playing Framework}
\label{all data create prompts}

The prompts for step 2, 3, and 4 are shown in \ref{prompt_details}.

\subsubsection{Annotators and Annotations}
\label{annotator-details}
\textbf{I. Annotation Recruiter Selection Process.}
We recruited annotators from a Chinese university and renowned enterprises through a three-stage selection process, ultimately choosing seven qualified annotators. The specific recruitment process includes three steps:
\begin{itemize}
    \item \textbf{Qualification Selection} Candidates must have a bachelor's degree or above, and background knowledge in NLP, and preference is given to those with annotation experience.
    \item \textbf{Trial Annotation} Candidates who pass the qualification selection are provided with uniform training. After the training, they perform trial annotations on a small-scale dataset, and the results are assessed for compliance with the annotation rules (for details, see \ref{annotation-guidelines-details}).
    \item \textbf{Diversity Selection} From those who pass the trial annotation, we select individuals from different university majors and professional departments to increase diversity.
\end{itemize}

\textbf{II. Measures to Ensure Dialogue Quality and Consistency}
Implement Unified Training by integrating a Training Manual, Meeting Clarifications, and Examples. Conduct pilot annotations to identify and resolve issues during the trial annotation period.
Timely communication, feedback, and discussion during the process, promptly supplement and update the training manual, and re-examine the annotated results.
Additionally, every instance was assigned to three annotators for cross-annotation validation with an inter-annotator agreement (IAA). 

\textbf{III. Compensation.}
On average, the annotation time was 10.3 minutes per sample, and the compensation was \$8 per hour.

\subsubsection{Annotation Guidelines}
\label{annotation-guidelines-details}
All annotators used the annotation tool for marking, and a screenshot of the annotation tool is shown in Figure \ref{annotation_tool_fig}. The complete annotation guideline includes numerous definitions and examples, with detailed formatting as shown above \ref{apdx: task_define_example}. Below is a brief explanation of the key annotation fields and important considerations.

\textbf{Step 1 annotation guidelines.}
Task 1 requires annotating the task definition. The task should come from various domains in the real world, and the task process can be described with core steps using SOP. The task can be completed in the form of a dialogue between the agent and the user.
Content to be annotated:

\begin{itemize}
    \item\textbf{Task Profile:} Provide the agent with relevant business knowledge about this dialogue task, which must include ``agent\_identity'', ``agent\_goal'' and ``success\_mark''. Additional necessary business knowledge can be supplemented.
    \item\textbf{User Profile:} Provide the agent with information about the user, who is the subject of this task dialogue, such as their name, etc.
    \item\textbf{Agent Action:} The dialogue actions that the agent can choose to facilitate the completion of the task.
    \item\textbf{User State:} The task status is achieved based on the information provided by the user during the dialogue process.
\end{itemize}

\textbf{Step 2 annotation guidelines.}
Task 2 requires the creation of SOP for the tasks defined in Task 1. The following should be annotated:
\begin{itemize}
    \item\textbf{SOP Vertexes:} Selected from agent actions and user states, used to define the core SOP diagram that needs to be followed.
    \item\textbf{SOP:} The interconnections between the nodes in the SOP diagram. It starts with ``Agent.Start'' and ends with ``Agent.PoliteEnd'', meaning that the leaf nodes have only one.
\end{itemize}

\textbf{Step 3 annotation guidelines.}
The dialogue path is a sequence of interactive actions generated by the LLM based on the SOP, representing a specific user completing a conversation in a particular setting. It is necessary to check whether the dialogue actions in the path follow the order in the core SOP. At the same time, for agent actions that are not part of the SOP, it is necessary to check whether they reflect the agent's proactivity, that is, whether the agent is striving to guide the conversation toward the task goal.

\textbf{Step 4 annotation guidelines.}
Annotators need to correct the dialogue content and dialogue actions (task status) to ensure they are consistent with the background knowledge provided for the task, comply with the constraints of the SOP, and adhere to common sense. Additionally, they should correct the grammatical correctness and fluency of the dialogue.

\subsubsection{List of Tasks in 32 Domains}
\label{domains detail}
All domains and tasks are shown in Table \ref{domain-task detail}.

\begin{table}[ht]
\small  
  \centering
    \resizebox{0.9\linewidth}{!}{ 
  \begin{tabular}{p{2.5cm}|p{6cm}} 
  \toprule
\textbf{Domains} & \textbf{Tasks}\\ \midrule
\textbf{scholarism}& conference invitation \\ \midrule
\textbf{courier}& delayed package handling \\ \midrule
\textbf{bank}& activate bank card, agent large transaction inquiry, financial product sales, golf invitation, loan followup \\ \midrule
\textbf{white goods}& repair appointment, installation appointment \\ \midrule
\textbf{shopping}& sams club member day invitation, redeem promotion, take out order \\ \midrule
\textbf{education}& online market \\ \midrule
\textbf{workplace}& apply for work card \\ \midrule
\textbf{photo studio}& photo appointment \\ \midrule
\textbf{hospital}& vaccine inform, appointment \\ \midrule
\textbf{airport}& check in \\ \midrule
\textbf{cosmetology}& product follow up, member day \\ \midrule
\textbf{household}& unblocking pipeline, moving appointment, property fee deposit, recycling appointment \\ \midrule
\textbf{restaurant}& private room booking, place a food order \\ \midrule
\textbf{cinema}& movie ticket purchase \\ \midrule
\textbf{pet}& complain consult, adoption facilitation \\ \midrule
\textbf{hotel}&check in\\ \midrule
\textbf{entertainment}&ktv complain consult\\ \midrule
\textbf{gym}&private tutoring, swimming pass promotion\\ \midrule
\textbf{car insurance}& sales promotion\\ \midrule
\textbf{community}& competition, lost and found\\ \midrule
\textbf{library}&borrow book\\ \midrule
\textbf{health}&blood pressure monitoring\\ \midrule
\textbf{telecom}&activate package, sim card upgrade promotion, broadband upgrade\\ \midrule
\textbf{domestic service}& complain\\ \midrule
\textbf{school}&home visit appointment, reissue student card, archive uery\\ \midrule
\textbf{tourism}& booking, hot spring promotion\\ \midrule
\textbf{real estate}& event invitation\\ \midrule
\textbf{internet}& broadband repair phone support\\ \midrule
\textbf{glasses}&fitting\\ \midrule
\textbf{computer}&repair appointment\\ \midrule
\textbf{account}&password recovery\\ \midrule
\textbf{survey}&membership reward\\ \bottomrule
  \end{tabular}
}
\caption{Domains and tasks Details.}    
  \label{domain-task detail} 
\end{table}

\subsection{Experiment Details}
\label{sec-exper-details}

\subsubsection{Offline Planner: Task 1}
The prompt for Direct Adjacency List (DAL) is shown in \ref{prompt_details}.
TCoT first-step: the prompt for LLM to describe each vertex and its child vertices in natural language is shown in \ref{prompt:tcot2}. TCoT second-step: the prompt for LLM to 
 translate the description into an adjacency list in JSON format is shown in \ref{prompt_details}

 \subsubsection{Online Planner: Task 2}
In MCTS, the number of generated actions $M$ is set to 3, depth limit $L$ is set to 8, the number of roll-outs $N$ is set to 3, and exploration weight $w$ is set to 1. To maintain a search scale similar to MCTS, in ToT, the number of generated actions $M$ is set to 3, depth limit $L$ is set to 8. 
The prompt for ToT sampling actions, generating responses, and predicting the user state is exactly the same as that for MCTS. 
The prompts for CoT, CoT+SOP, MCTS, and ToT are shown in \ref{prompt_details}. 

\subsubsection{Supervised Fine-Tuning}
All open-source models' experiments were completed on 4 Nvidia A800 GPUs. For SFT, we fine-tuned all parameters using 5 epochs, 50 warm steps, 128 batch size, and 1e-6 learning rate. 

\subsubsection{Generation Parameters}
In the experiment of task 1, the temperature was uniformly set to 0.1 and the top-p was set to 0.1. For task 2, the temperature was uniformly set to 1 and the top-p was set to 0.95. 

\subsection{Human Evaluation Details}
\label{manually_evaluate}
We continue to employ the annotators recruited for dataset construction to complete the evaluation of different models in dialogue tests. 
Let $H$ denote the annotated preceding dialogue context, and let $r$ represent the model's agent response to $H$.
The specific evaluations include:
%
%

\textbf{Controllability Accuracy} If the content of $r$ originates from the SOP, then $r$ is added to a controllable response set $R_{\text{ctrl}} $. Furthermore, if $r$ correctly adheres to the SOP procedural constraints, then $r$ is added to a correctly controllable response set $R_{\text{ctrl\_right}} $. Finally, the controllability accuracy $acc_{\text{ctrl}} $ is defined as:
\begin{align}
acc_{\text{ctrl}} = \frac{\mathrm{count}(R_{\text{ctrl\_right}})}{\mathrm{count}(R_{\text{ctrl}})}
\end{align}

\textbf{Proactivity Accuracy} If the content of $r$ originates not from the SOP, then $r$ is added to a proactive response set $R_{\text{pact}} $. Furthermore, if $r$ responds correctly to the user and effectively guides the conversation towards achieving the goal, then $r$ is added to a correctly proactive response set $R_{\text{pact\_right}} $. Finally, the proactivity accuracy $acc_{\text{pact}} $ is defined as:
\begin{align}
acc_{\text{pact}} = \frac{\mathrm{count}(R_{\text{pact\_right}})}{\mathrm{count}(R_{\text{pact}})}
\end{align}

\textbf{Knowledge Accuracy} If $r$ contains business or user information from the task, it will be added to the knowledge response set $R_{know}$. In addition, if $r$ contains accurate information, free of errors or hallucinations, then $r$ will be added to the correct knowledge response set $(r_{\text{know\_right}})$.Finally, the knowledge accuracy $acc_{\text{know}} $ is defined as:
\begin{align}
acc_{\text{know}} = \frac{\mathrm{count}(R_{\text{know\_right}})}{\mathrm{count}(R_{\text{know}})}
\end{align}

\textbf{Goal Success Rate} If the model's responses in each turn contain no controllability errors and knowledge errors, and the dialogue ultimately reaches the goal position, the dialogue is considered goal successful. The goal success rate is the proportion of successful dialogues to the total number of dialogues.

\textbf{Logical Coherence Score} We evaluate the logical coherence of the model by breaking down the scoring points across multiple dimensions. 
\begin{itemize}
    \item Response Logic: Whether the model misunderstands the user's intent and responds with content that is completely unrelated or incorrect to the intent. If there is an error, score 0; otherwise, score 0.4.
    \item Context Consistency: Whether the model's response content contradicts the dialogue history. If there is a contradiction, score 0; otherwise, score 0.3.
    \item Common Sense: If the model's response contradicts common sense, score 0 else 0.3.
\end{itemize}



\subsection{Example of Dialogue}
\begin{lstlisting}[style=dialogue, caption=An Example of Dialogue, basicstyle={\fontsize{7.5pt}{8.5pt}\selectfont}]

Agent Action: Greet
Agent Response: Hi, this is the customer service center of the ** Bank Credit Card Department.

User Response: Hi!
User State: Greet

Agent Action: VerifyIdentity
Agent Response: Hi, am I speaking with Ms. Hao **?

User Response: Yes, that's right.
User State: Cardholder

Agent Action: IntroduceActivationActivities.
Agent Response: Ms. Hao, thank you for applying for our platinum credit card. To reward our new customers, we currently have an activation promotion that offers a 12\% discount on fuel for one year. Once you activate your credit card, you can enjoy this benefit.

User Response: That sounds great, but I've never used a credit card before and I'm a bit worried it might be troublesome.
User State: WorryAndDoubt

Agent Action: RelieveDoubts
Agent Response: Ms. Hao, you don't have to worry at all. Activating a credit card is very simple. You can choose to handle it at the bank counter, or you can activate it through the mobile banking APP or by calling our credit card customer service hotline for self-activation. The whole process is very convenient and fast.

User Response: Oh, I see. That's good to know.
User State:  HabitualResponseAndContinue

Agent Action:  AttemptPersuasion 
Agent Response: Yes, and after activation, you can also set a transaction password to ensure the security of your card usage. Our platinum credit card is free of annual fees for the first year, and you can waive the annual fee by swiping the card three times a year. This is not only safe and convenient but also comes with many promotional activities.

User Response: That sounds like a great deal, please activate it for me.
User State: Activate

Agent Action: SetPassword
Agent Response: Thank you for your trust. To ensure the security of your card transactions, please set a 6-digit transaction password.

User Response: Sure, my transaction password is ******.
User State: Success

Agent Action: Thank
Agent Response: Ms. Hao, your credit card has been successfully activated, and the transaction password has been set. You can now use your credit card normally.

User Response: Thank you, it's really convenient.
User State: Thank

Agent Action: Goodbye
Agent Response: You're welcome, Ms. Hao. If you have any questions, you can call our customer service hotline at any time. I wish you a happy life, goodbye!

User Response: Okay, goodbye!
User State: Goodbye

\end{lstlisting}

\subsection{Prompt Details}
\label{prompt_details}
\begin{tcolorbox}[
    colframe=grenn_2,         
    colback=white,           
    coltitle=black,          
    colbacktitle=grenn_2,       
    rounded corners,
    arc=4mm, 
    enhanced,                
    boxrule=0.8mm,           
    frame style={solid},     
    fonttitle=\bfseries,     
    fontupper=\small,
    title={Prompting LLM to predict SOP's Direct Adjacency List (DAL).} 
]
You are a professional business manager, having accumulated extensive knowledge of business processes and being adept at conversing with clients to complete business tasks. \\

Now, please create the standard operating procedure (SOP) diagram for this task based on the following ``Task Profile''. All vertexes of this SOP diagram are as follows ``SOP Vertexes'', and you need to output the list of adjacent for each vertex. \\

Please strictly follow the requirements below: \\
1. The output format should be in the form of a JSON string array. \\
2. The JSON fields must include all nodes. If a vertex has no adjacent vertexes, represent it with []. \\

\textbf{Task Profile} 
\begin{minted}[bgcolor=gray!5, fontsize=\tiny, breaklines, frame=none]{json}
{ 
    "agent_identity": "** Bank Credit Card Center Customer Service", 
    "task_goal": "To persuade the user to explicitly agree to the credit card", 
    "success\_mark": [ 
        "User.ClearAgreement"
        ], 
    "...":"...", 
    "other_knowledge": "Credit Card Opening Guide I. Card Activation ..." 
}
\end{minted}

\textbf{SOP Vertexes }
\begin{minted}[bgcolor=gray!5, fontsize=\tiny, breaklines, frame=none]{json}
[ 
     "Agent.Start", 
     "Agent.VerifyIdentity", 
     "...", 
     "Agent.PoliteEnd", 
     "Agent.NotifyActivationSuccess" 
]
\end{minted}

The adjacency list of all vertexes: \\
\noindent\rule{\linewidth}{0.4pt}
\textbf{Generation Output:}
\begin{minted}[bgcolor=gray!5, fontsize=\tiny, breaklines, frame=none]{json}
{ 
 "Agent.Start":[
    "Agent.VerifyIdentity"
    ], 
 "Agent.VerifyIdentity":[ 
     "User.Cardholder",
     "User.NonCardholder"
     ], 
 "...":[
    "..."
    ], 
 "User.ClearRejection":[
    "Agent.PoliteEnd"
    ], 
 "Agent.PoliteEnd":[], 
 "Agent.NotifyActivationSuccess":[
    "Agent.PoliteEnd"
    ]
}
\end{minted}
\end{tcolorbox}


\begin{tcolorbox}[
    colframe=grenn_2,         
    colback=white,           
    coltitle=black,          
    colbacktitle=grenn_2,       
    rounded corners,
    arc=4mm, 
    enhanced,                
    boxrule=0.8mm,           
    frame style={solid},     
    fonttitle=\bfseries,     
    fontupper=\small,
    title={The prompt of TCoT to translate the description into an adjacency list.\label{prompt:tcot2}} 
]
You are a seasoned business representative, having accumulated a wealth of business process knowledge, and are adept at conversing with customers to complete transactions. Now please formulate the Standard Operating Procedure (SOP) diagram for this task. \\

You can refer to the following ``Task Knowledge'', but it may not be comprehensive or accurate, and you do not need to strictly follow it. All nodes of the SOP diagram are as follows ``SOP Diagram Nodes'', and you need to output the list of adjacent nodes for each node, that is, the adjacency list of the entire SOP diagram. \\

\textbf{Hint} \\

1. The SOP diagram starts with `Agent.Start' and ends with `Agent.PoliteEnd'. \\
2. The subsequent nodes of the ``User'' node are usually ``Agent'' nodes, and the subsequent nodes of the ``Agent'' node mostly require a response from the ``User'' node. For some actions that do not require a response, the subsequent node of ``Agent'' can also be ``Agent''. \\
3. If there is a `VerifyIdentity' node in the task, it indicates that only the person themselves can continue, and the conversation ends if it is not the person themselves. \\
4. The JSON fields must include all nodes. \\
5. Please strictly follow output format, for example: \\

\begin{minted}[bgcolor=gray!5, fontsize=\tiny, breaklines, frame=none]{json}
{ 
    "Agent.Start": [
        "Agent.Node1"
        ], 
    "Agent.Node1": [
        "User.Node1", 
        "User.Node2"
        ], 
    "...": ["..."]
}
\end{minted}

\textbf{Task Knowledge} \\
After `Agent.Start', the first step should be `Agent.VerifyIdentity'. After confirming the identity, it could be `User.IsThemselves' or `User.NotThemselves'. If not the person, the agent will `Agent.PoliteEnd'. If the user is the person, then the agent will introduce the event, that is `Agent.InviteToGolfExperienceEvent'...\\

\textbf{SOP Diagram Nodes}
\text{[} ``Agent.Start'',  ``Agent.VerifyIdentity'',  ... \text{]} \\

Adjacency list for all nodes:\\

\noindent\rule{\linewidth}{0.1pt}
\textbf{Generation Output:}
\begin{minted}[bgcolor=gray!5, fontsize=\tiny, breaklines, frame=none]{json}
{
    "Agent.Start":[
        "Agent.VerifyIdentity"
        ],
    "...":["..."]
}
\end{minted}

\end{tcolorbox}
\begin{tcolorbox}[
    colframe=grenn_2,         
    colback=white,           
    coltitle=black,          
    colbacktitle=grenn_2,       
    rounded corners,
    arc=4mm, 
    enhanced,                
    boxrule=0.8mm,           
    frame style={solid},     
    fonttitle=\bfseries,     
    fontupper=\small,
    title={The prompt of TCoT to describe the adjacency relationship between vertexes in natural language.} 
]
 You are a professional business manager, having accumulated a wealth of business process knowledge and being adept at conversing with customers to complete transactions. \\

Now, based on the vertexes of the Standard Operating Procedure (SOP) diagram, please analyze all the vertexes. Write a passage analyzing the entire process of interaction and communication between the `User' and `Agent', covering all vertexes and all possible scenarios, such as what happens if the user `Agree' or `Disagree'. \\

\textbf{Here is an example:} \\
\textbf{SOP vertexes: }\\

\text{[}`Agent.Start', `Agent.VerifyIdentity', `User.IsThemselves', `Agent.IntroduceGiftDelivery', `User.NotThemselves', `User.ClearAgreement', `Agent.AskForDeliveryAddress', `User.ProvideDeliveryAddress', `User.DoNotProvideDeliveryAddress', `User.ClearRejection', `Agent.PoliteEnd', `Agent.NotifyRegistrationSuccess'\text{]} \\

\textbf{Start analysis:}\\
After `Agent.Start', the first step should be `Agent.VerifyIdentity'. After confirming the identity, it could be `User.IsThemselves' or `User.NotThemselves'. If not the person, the agent will `Agent.PoliteEnd'. If the user is the person, the agent will introduce, that is `Agent.IntroduceGiftDelivery'. If `User.ClearAgreement', the process can continue. Since it is a gift delivery, after the user agrees, `Agent.AskForDeliveryAddress', if `User.ProvideDeliveryAddress' then the agent `Agent.NotifyRegistrationSuccess', if `User.DoNotProvideDeliveryAddress', the agent may ask again `Agent.AskForDeliveryAddress', or directly `Agent.PoliteEnd'. After `Agent.NotifyRegistrationSuccess', it can `Agent.PoliteEnd'. \\

\textbf{SOP vertexes: }\\
\text{[} ``Agent.Start'', ``Agent.VerifyIdentity'', ...,``Agent.PoliteEnd''\text{]} \\

\textbf{Start analysis:}\\

\noindent\rule{\linewidth}{0.1pt}
\textbf{Generation Output:}\\
After `Agent.Start', the first step should be `Agent.VerifyIdentity'. After confirming the identity, it could be `User.IsThemselves' or `User.NotThemselves'. If not the person, the agent will `Agent.PoliteEnd'. If the user is the person, then the agent will introduce the event, that is `Agent.InviteToGolfExperienceEvent'. If ...\\ \\ \\

\end{tcolorbox}

\begin{tcolorbox}[
    breakable,
    colframe=grenn_2,         
    colback=white,           
    coltitle=black,          
    colbacktitle=grenn_2,       
    rounded corners,
    arc=4mm, 
    enhanced,                
    boxrule=0.8mm,           
    frame style={solid},     
    fonttitle=\bfseries,     
    fontupper=\small,
    title={Prompting LLM to generate full dialogue paths.} 
]
You are a professional business manager, assisting customers in handling their business through dialogue.

Based on the provided ``task\_profile'', ``user\_profile'' and ``sop\_adjacency\_list'', Please insert more ``user\_state'' and ``agent\_actions'' into the above ``Main Dialogue Path'' to enrich the possible reactions of the user and the corresponding decision-making actions that the agent takes in response to the user's state, forming a full dialogue path. 

Below is the related information: \\
\begin{minted}[bgcolor=gray!5, fontsize=\tiny, breaklines, frame=none]{json}
{
    "task_profile": {"...":"..."},
    "user_profile": {"...":"..."},
    "sop_adjacency_list": {"...":["..."]},
    "agent_action": [
        "Agent.VerifyIdentity", 
        "..."
    ], 
    "user_state": [
        "User.Cardholder", 
        "..."
    ]
}
\end{minted}

\# Main Dialogue Path: \\
\text{[ ``Agent.VerifyIdentity'', ``User.Cardholder''... ]} \\

Please follow the requirements below strictly: \\

1. Full dialogue paths should start with ``Agent.Greeting'' and end with ``User.Ending''. \\
2. Require ``Agent Action'' and ``User State'' to be spoken alternately, with each occurrence representing a round of dialogue. Please add a ``--'' after each round of dialogue to separate the previous and next rounds. For example,  [``Agent.Greeting'', ``User.Greeting'', ``--'', ``Agent.IntroduceActivity'',...]. \\
3. Don't delete nodes or adjust the nodes' order in the ``Main Dialogue Path''. \\
4. Please only insert 2 to 5 reasonable rounds of dialogue. \\
5. The inserted nodes can only be selected from the ``user\_state'' and ``agent\_action'' defined in the above text, and do not create new names. \\
6. When the preceding node comes from ``sop'', please try to choose one from the subsequent nodes of the current node in the adjacency list. \\
7. The output format should be a JSON string array. \\

Supplemented full dialogue paths in JSON format: 

\noindent\rule{\linewidth}{0.4pt}
\textbf{Generation Output:}
\begin{minted}[bgcolor=gray!5, fontsize=\tiny, breaklines, frame=none]{json}
[
    "Agent.Greeting", 
    "User.Greeting", 
    "--", 
    "Agent.VerifyIdentity",
    "User.Cardholder",
    "--", 
    "Agent.IntroduceActivity"
    "..."
]
\end{minted}

\end{tcolorbox}

\begin{tcolorbox}[
    breakable,
    colframe=grenn_2,         
    colback=white,           
    coltitle=black,          
    colbacktitle=grenn_2,       
    rounded corners,
    arc=4mm, 
    enhanced,                
    boxrule=0.8mm,           
    frame style={solid},     
    fonttitle=\bfseries,     
    fontupper=\small,
    title={Prompting LLM to generate dialogue according to full dialogue paths.} 
]
You are a professional business manager, assisting customers with their business through dialogues. \\

Now, due to the need for training new business representatives, you are asked to act as a scriptwriter. Based on the ``task\_profile'' and ``user\_profile'' provided below, and the ``Full Dialogue Path'' between the agent and the user for this task, write dialogue scripts that fit the roles of both parties. \\
Here are the task-related information: \\

\begin{minted}[bgcolor=gray!5, fontsize=\tiny, breaklines, frame=none]{json}
{
    "task_profile":{ 
        "agent_identity": "** Bank Credit Card Center Customer Service", 
        "...", 
        "other_knowledge": "Credit Card Activation Guide I. Card Activation ..."
    }, 
    "user_profile": {
        "Name": "**", 
        "Gender": "Male", 
        "...", 
        "CardID":"****"
    }
}
\end{minted}

\textbf{Full Dialogue Paths:}\\
\text{[}``Agent.Greeting'', ``User.Greeting'', ``Agent.VerifyIdentity'', 
``User.Cardholder'', ... \text{]}\\

Please write dialogue scripts for the ``Full Dialogue Paths'' based on the task-related information, forming a complete conversation. \\

Please strictly follow the requirements below:\\

1.Please directly append the ``script'' you have written after the name of each node in the ``Complete Dialogue Path'', separated by the ``|'' symbol. For example: ``User.Consider|I understand, I will think about it.''\\

2.Do not add, delete, or change any original node.\\

3.The complete output format should be consistent with the ``Complete Dialogue Path'', that is, output as a JSON array of strings, such as [``Agent.PoliteEnd|Thank you for your cooperation, I wish you a happy life!'', ``User.Ending|Okay, goodbye!'']\\

Please output the ``Complete Dialogue Path'' after adding the dialogue scripts:\\

\noindent\rule{\linewidth}{0.1pt}
\textbf{Generation Output:}
\begin{minted}[bgcolor=gray!5, fontsize=\tiny, breaklines, frame=none]{json}
[
    "Agent.Greeting|Hello, this is the credit card center of ** Bank.", 
    "User.Greeting|Hello",
    "Agent.VerifyIdentity|Hi, may I ask if you are Ms. **?",
    "User.Cardholder| Yes, this is ** speaking. ",
    "Agent.IntroduceActivationActivity|That's correct, Mr.* we have noticed that your platinum credit card has not been activated yet. To express our gratitude for your support, we currently have a promotion where you can enjoy a 12\% discount on fuel for one year just by activating your card.", 
    "..."
] 
\end{minted}

\end{tcolorbox}

\begin{tcolorbox}[
    breakable,
    valign=top,
    colframe=grenn_2,         
    colback=white,           
    coltitle=black,          
    colbacktitle=grenn_2,       
    rounded corners,
    arc=4mm, 
    enhanced,                
    boxrule=0.8mm,           
    frame style={solid},     
    fonttitle=\bfseries,     
    fontupper=\small,
    title={The prompt for sampling candidate agent action to expand in SGM.} 
]
Given the following ``user\_profile'', ``task\_profile'' and ``Dialogue Context''. Please select the next round of agent action from the ``Optional Agent Actions'' for the agent that can best encourage the user to complete this dialogue task. When the task is completed, please actively choose ``Agent.PoliteEnd'' to end the conversation. The name of the agent action must be exactly the same as those in the candidate list. \\

\begin{minted}[bgcolor=gray!5, fontsize=\tiny, breaklines, frame=none]{json}
{
    "task_profile":{ 
        "agent_identity": "** Bank Credit Card Center Customer Service", 
        "...", 
        "other_knowledge": "Credit Card Activation Guide I. Card Activation ..."
    }, 
    "user_profile": {
        "Name": "**", 
        "Gender": "Male", 
        "...", 
        "CardID":"****"
    }
}
\end{minted}

\textbf{Optional Agent Actions:}  \\
\text{[}``Start'', ``VerifyIdentity'', ``IntroduceActivationActivity'', ...,``RelieveDoubts'', ``AttemptPersuasion'', ``Chat'', ``OtherActions''\text{]} \\

\textbf{Dialogue Context:}  \\
Agent Action: Greeting \\
Agent Response: Hello, this is the credit card center of
** Bank. \\
User Response: Hello! \\
User State: Greeting \\
Agent Action: VerifyIdentity \\
Agent Response: Hi, may I ask if you are Ms. **?
User Response: Yes, this is ** speaking. \\

\textbf{Dialogue SOP}:  \\
Based on the standard procedure for this task, please prioritize the following agent actions: \\
\text{[}``IntroduceActivationActivity''\text{]} \\

Start with ``Analysis:'' and deduct step by step the best agent actions that should be used by the subsequent dialogue agent, concluding with an output similar to ``Therefore, the best agent action is: Greeting'', with a colon required before the name of the agent action. \\

\noindent\rule{\linewidth}{0.1pt}
\textbf{Generation Output:}\\

Analysis: According to the previous conversation, the user confirms their identity, indicating that the user is the person the agent needs to verify. Following the task process prompts, the next step is to invite them to participate in the activation activity, which aligns with the conversation task logic. \\

Therefore, the best agent action is: IntroduceActivationActivity\\
\end{tcolorbox}

\begin{tcolorbox}[
    breakable,
        valign=top,
    colframe=grenn_2,         
    colback=white,           
    coltitle=black,          
    colbacktitle=grenn_2,       
    rounded corners,
    arc=4mm, 
    enhanced,                
    boxrule=0.8mm,           
    frame style={solid},     
    fonttitle=\bfseries,     
    fontupper=\small,
    title={The prompt generates an agent response for MCTS's simulation.} 
]
Given the ``user\_profile'', ``task\_profile'', and ``Dialogue Context'' for the dialogue task below, please generate an appropriate agent response by following the chosen agent action. Start your response with a new line and ``Agent Response:''. \\
\begin{minted}[bgcolor=gray!5, fontsize=\tiny, breaklines, frame=none]{json}
{
    "task_profile":{ 
        "agent_identity": "** Bank Credit Card Center Customer Service", 
        "...", 
        "other_knowledge": "Credit Card Activation Guide I. Card Activation ..."
    }, 
    "user_profile": {
        "Name": "**", 
        "Gender": "Male", 
        "...", 
        "CardID":"****"
    }
}
\end{minted}

\textbf{Dialogue Context:}  \\
Agent Action: Greeting \\
Agent Response: Hello, this is the credit card center of
** Bank. \\
User Response: Hello! \\
User State: Greeting \\
Agent Action: VerifyIdentity \\
Agent Response: Hi, may I ask if you are Ms. **?
User Response: Yes, this is ** speaking. \\
User State: Cardholder \\
Agent Action: IntroduceActivationActivity

\noindent\rule{\linewidth}{0.1pt}
\textbf{Generation Output:}\\
Agent Response: Hi Mr. **, this is the credit card center of
** Bank. I noticed you haven't activated your Platinum credit card yet, and I wanted to thank you for being a valued customer. Right now, we're running a special promotion where you can get a 12\% discount on fuel for an entire year just by activating your card. It's a great way to save on your everyday expenses. Would you like me to help you get started with the activation today? 
\\
\\
\\
\\
\\
\\
\\
\\
\\
\\
\\
\\
\\
\\
\\
\\
\\
\\
\\
\\
\\
\\
\end{tcolorbox}


\begin{tcolorbox}[
    colframe=grenn_2,         
    colback=white,           
    coltitle=black,          
    colbacktitle=grenn_2,       
    rounded corners,
    arc=4mm, 
    enhanced,                
    boxrule=0.8mm,           
    frame style={solid},     
    fonttitle=\bfseries,     
    fontupper=\small,
    title={The prompt for MCTS to get a reward score.} 
]
You are a business expert manager at a company in the financial sector. The company's business agent is contacting the customer below to handle affairs, see ``Dialogue Context'' for the content of the conversation. Now, please carefully read follow information and judge whether the agent's chosen ``Agent action'' conforms to the task process (sop) or is conducive to guiding the user toward the direction of task success. If it meets the requirements, it is 1; otherwise, it is 0. \\

\begin{minted}[bgcolor=gray!5, fontsize=\tiny, breaklines, frame=none]{json}
{
    "task_profile":{ 
        "agent_identity": "** Bank Credit Card Center Customer Service", 
        "...", 
        "other_knowledge": "Credit Card Activation Guide I. Card Activation ..."
    }, 
    "user_profile": {
        "Name": "**", 
        "Gender": "Male", 
        "...", 
        "CardID":"****"
    },
    "sop_adjacency_list": {"...":["..."]},
    "agent_action": [
        "Agent.VerifyIdentity", 
        "..."
    ]
}
\end{minted}

\textbf{Dialogue Context:}  \\
Agent Action: Greeting \\
Agent Response: Hello, this is the credit card center of
** Bank. \\
User Response: Hello! \\
User State: Greeting \\
Agent Action: VerifyIdentity \\
Agent Response: Hi, may I ask if you are Ms. **?
User Response: Yes, this is ** speaking. \\
User State: Cardholder \\
Agent Action: IntroduceActivationActivity\\

Start your analysis with ``Analysis:'' and after your conclusion, summarize with ``Therefore, the answer is: 1'' or ``Therefore, the answer is: 0''. \\

\noindent\rule{\linewidth}{0.1pt}
\textbf{Generation Output:} \\
Analysis: According to the task knowledge, the objective of the agent's task is to invite users to participate in credit card activation event. The agent action chosen is IntroduceActivationActivity, which aligns with the task objective. Therefore, it is consistent with the above handling process and the best action within the optional agent actions. \\

Therefore, the answer is: 1\\

\end{tcolorbox}


\begin{tcolorbox}[
    colframe=grenn_2,         
    colback=white,           
    coltitle=black,          
    colbacktitle=grenn_2,       
    rounded corners,
    arc=4mm, 
    enhanced,                
    boxrule=0.8mm,           
    frame style={solid},     
    fonttitle=\bfseries,     
    fontupper=\small,
    title={Prompting LLM to generate user state.} 
]
You are an agent specializing in lifestyle services. Given the following ``user\_profile'', ``task\_profile', and ``Dialogue Context'', please select an option from the ``Optional User State'' that best reflects the user's current task status in the dialogue context. \\

Start with ``User State:'' and the state name must be exactly the same as one in the list. \\

\begin{minted}[bgcolor=gray!5, fontsize=\tiny, breaklines, frame=none]{json}
{
    "task_profile":{ 
        "agent_identity": "** Bank Credit Card Center Customer Service", 
        "...", 
        "other_knowledge": "Credit Card Activation Guide I. Card Activation ..."
    }, 
    "user_profile": {
        "Name": "**", 
        "Gender": "Male", 
        "...", 
        "CardID":"****"
    }
}
\end{minted}

\textbf{Dialogue Context:}
Agent Action: Greeting \\
Agent Response: Hello, this is the credit card center
of ** Bank. \\
User Response: Hello! \\
User State: Greeting \\
Agent Action: VerifyIdentity \\
Agent Response: Hi, may I ask if you are Ms. **? \\
User Response: Yes, this is ** speaking. \\

\textbf{Optional User State:} \\
\text{[}"Greet", "Cardholder","..."\text{]}

\noindent\rule{\linewidth}{0.1pt}
\textbf{Generation Output:}\\
User State: Cardholder 
\\
\\
\\
\\
\\
\\
\\
\\
\\
\\
\\
\\
\\
\\
\\
\\
\\
\\
\\
\\
\\

\end{tcolorbox}


\begin{tcolorbox}[
    colframe=grenn_2,         
    colback=white,           
    coltitle=black,          
    colbacktitle=grenn_2,       
    rounded corners,
    arc=4mm, 
    enhanced,                
    boxrule=0.8mm,           
    frame style={solid},     
    fonttitle=\bfseries,     
    fontupper=\small,
    title={The prompt for CoT.} 
]
You are a business expert agent in the financial field. Given the following ``user\_profile'', ``task\_profile', and ``Dialogue Context'', please output the following three items \textbf{step by step}: \\
1. Please select an option from ``Optional User State'' that best reflects the user's current task status in the dialogue above, and start with ``User State:''. Its name must match exactly with one from the optional list. \\
2. Please select an option from  ``Optional Agent Action'' that best reflects the agent's next round of action that can prompt the user to complete the dialogue task. Start with a new line and ``Agent Action:''. Its name must match exactly with the one on the candidate list. \\
3. Please generate an appropriate Agent Response combined with the agent's chosen dialogue action. Start with a new line and ``Agent Response:'' \\

\begin{minted}[bgcolor=gray!5, fontsize=\tiny, breaklines, frame=none]{json}
{
    "task_profile": {"...":"..."},
    "user_profile": {"...":"..."},
    "agent_action": [
        "Agent.VerifyIdentity",
        "..."
    ],
    "user_state": [
        "User.Cardholder",
        "..."
    ]
}
\end{minted}

\textbf{Dialogue Context:}\\
Agent Action: Greeting \\
Agent Response: Hello, this is the credit card center
of ** Bank. \\
User Response: Hello! \\
\\
\\
\\
\\
\\
\\
\\

\noindent\rule{\linewidth}{0.1pt}
\textbf{Generation Output:}\\
User State: Greeting \\
Agent Action: VerifyIdentity \\
Agent Response: Hi, may I ask if you are Ms. **?
\end{tcolorbox}


\begin{tcolorbox}[
    colframe=grenn_2,         
    colback=white,           
    coltitle=black,          
    colbacktitle=grenn_2,       
    rounded corners,
    arc=4mm, 
    enhanced,                
    boxrule=0.8mm,           
    frame style={solid},     
    fonttitle=\bfseries,     
    fontupper=\small,
    title={The prompt for CoT+SOP.} 
]
You are a business expert agent in the financial field. Given the following ``user\_profile'', ``task\_profile', and ``Dialogue Context'', please output the following three items \textbf{step by step}: \\
1. Please select an option from ``Optional User State'' that best reflects the user's current task status in the dialogue above, and start with ``User State:''. Its name must match exactly with one from the optional list. \\
2. Please select an option from  ``Optional Agent Action'' that best reflects the agent's next round of action that can prompt the user to complete the dialogue task. Start with a new line and ``Agent Action:''. Its name must match exactly with the one on the candidate list. \\
3. Please generate an appropriate Agent Response combined with the agent's chosen dialogue action. Start with a new line and ``Agent Response:'' \\

\begin{minted}[bgcolor=gray!5, fontsize=\tiny, breaklines, frame=none]{json}
{
    "task_profile": {"...":"..."},
    "user_profile": {"...":"..."},
    "agent_action": [
        "Agent.VerifyIdentity",
        "..."
    ],
    "user_state": [
        "User.Cardholder",
        "..."
    ]
}
\end{minted}

\textbf{Dialogue Context:}\\
Agent Action: Greeting \\
Agent Response: Hello, this is the credit card center
of ** Bank. \\
User Response: Hello! \\

\textbf{Dialogue SOP}:\\
Based on the standard process of handling this task and the dialogue context, please prioritize the following agent action that meets the Standard Operating Procedures (SOP): \\
\text{[}``VerifyIdentity''\text{]} \\
        
\noindent\rule{\linewidth}{0.1pt}
\textbf{Generation Output:}\\
User State: Greeting \\
Agent Action: VerifyIdentity \\
Agent Response: Hi, may I ask if you are Ms. **?

\end{tcolorbox}

\onecolumn

\subsection{Algorithm Details}
Algorithm \ref{algorithm mcts} presents the pseudocode for a Monte Carlo Tree Search (MCTS) planning algorithm designed for large language model (LLM) dialogue agents. The algorithm aims to improve dialogue policy quality by effectively exploring and selecting dialogue actions through simulation and backpropagation mechanisms.

\label{sec-algorithm}
\begin{algorithm*}[tb]
\fontsize{8pt}{9pt}\selectfont
\caption{ Online Planning for LLM Dialogue Agents with MCTS}
\label{algorithm mcts}
\begin{algorithmic}[1]
\Require Initial dialogue state $d_0$, state transition probability function $p_{\theta}$, reward function $r_{\theta}$, action generator $a_{\phi}$
\Require Number of expand actions $M$, depth limit $L$, number of roll-outs $N$, and exploration weight $w$
\State Initialize memory of actions $A : D \to A$, children $c : D \times A \to D$ and rewards $r : D \times A \to \mathbb{R}$
\State Initialize the state-action value function $Q : D \times A \to \mathbb{R}$ and visit counter $N : D \to \mathbb{N}$
\For {$k \leftarrow 0, \dots, N - 1$}
    \State $t \leftarrow 0$
    \While {$N(d_t) > 0$} \Comment\{Selection\}
        \State $N(d_t) \leftarrow N(d_t) + 1$
        \State $a_t \leftarrow \text{arg max}_{p \in A(d_t)} \left[ Q(d_t, p) + w \sqrt{\frac{\ln N(d_t)}{N(c(d_t, p))}} \right]$
        \State $r_t = r(d_t, a_t), d_{t+1} \leftarrow c(d_t, a_t)$
        \State $t \leftarrow t + 1$
    \EndWhile
    \While {$d_t$ is not a terminal state $\wedge t \leq L$} \Comment{Expansion}
        \For {$i \leftarrow 1, \dots, M$}
            \State Sample $a_t^{(i)} \sim a_{\phi}(p | d_t), d_{t+1}^{(i)} \sim p_{\theta}(d_t, a_t^{(i)}), r_t^{(i)} \sim r_{\theta}(d_t, a_t^{(i)})$
            \State Update $A(d_t) \leftarrow \left\{ a_t^{(i)} \right\}_{i=1}^{d}, c(d_t, a_t^{(i)}) \leftarrow d_{t+1}^{(i)}, r(d_t, a_t) \leftarrow r_t^{(i)}$
        \EndFor
        \State $a_{t+1} \leftarrow \text{arg max}_{a \in A(d_t)} r(d_t, a_t)$ \Comment{Simulation}
        \State $r_t \leftarrow r(d_t, a_t), d_{t+1} \leftarrow c(d_t, a_t)$
        \State $t \leftarrow t + 1$
    \EndWhile 
    \For {$t' \leftarrow t, \dots, 0$} \Comment{Back propagation}
        \State Update $Q(d_{t'}, a_{t'})$ with $\{ r_{t'}, r_{t'+1}, \dots, r_t \}$
    \EndFor
\EndFor
\end{algorithmic}
\end{algorithm*}